\documentclass[11pt,a4paper,oldfontcommands]{memoir}
\usepackage[utf8]{inputenc}
\usepackage[T1]{fontenc}
\usepackage{microtype}
\usepackage[dvips]{graphicx}
\usepackage{xcolor}
\usepackage{times}
\usepackage{pbox}
\usepackage{algorithmic}
\usepackage{algorithm}
\usepackage{amsmath}
\usepackage{caption}
\usepackage{subfig}

\usepackage[
breaklinks=true,colorlinks=true,
linkcolor=black,urlcolor=black,citecolor=black,% PRINT
bookmarks=true,bookmarksopenlevel=2]{hyperref}

\usepackage{geometry}
\OnehalfSpacing
\chapterstyle{bianchi}
\setsecheadstyle{\Large\bfseries\sffamily\raggedright}
\setsubsecheadstyle{\large\bfseries\sffamily\raggedright}
\setsubsubsecheadstyle{\bfseries\sffamily\raggedright}

\makepagestyle{plain}
\makeevenfoot{plain}{\thepage}{}{}
\makeoddfoot{plain}{}{}{\thepage}
\makeevenhead{plain}{}{}{}
\makeoddhead{plain}{}{}{}

\maxsecnumdepth{subsection} % chapters, sections, and subsections are numbered
\maxtocdepth{subsection} % chapters, sections, and subsections are in the Table of Contents

\DeclareMathOperator*{\argmin}{argmin}
\DeclareMathOperator*{\argmax}{argmax}

\begin{document}

%%%---%%%---%%%---%%%---%%%---%%%---%%%---%%%---%%%---%%%---%%%---%%%---%%%
%   TITLEPAGE
%
%   due to variety of titlepage schemes it is probably better to make titlepage manually
%
%%%---%%%---%%%---%%%---%%%---%%%---%%%---%%%---%%%---%%%---%%%---%%%---%%%
\thispagestyle{empty}

{%%%
\sffamily
\centering
\Large

~\vspace{\fill}

{\huge 
Minimally Supervised Feature Selection for Classification
}

\vspace{2.5cm}

{\LARGE
Radu Alexandra Maria
}

\vspace{3.5cm}

Master Thesis\\[1em]
%in the\\[1em]

University Politehnica of Bucharest\\
Faculty of Automatic Control and Computers \\
Computer Science and Engineering Department \\
\vspace{3.5cm}

Supervisor: Associate Prof. Dr. Marius Leordeanu

\vspace{\fill}

July 2015

%%%
}%%%

\cleardoublepage
%%%---%%%---%%%---%%%---%%%---%%%---%%%---%%%---%%%---%%%---%%%---%%%---%%%
%%%---%%%---%%%---%%%---%%%---%%%---%%%---%%%---%%%---%%%---%%%---%%%---%%%

\tableofcontents*

\clearpage

%%%---%%%---%%%---%%%---%%%---%%%---%%%---%%%---%%%---%%%---%%%---%%%---%%%
%%%---%%%---%%%---%%%---%%%---%%%---%%%---%%%---%%%---%%%---%%%---%%%---%%%
\chapter*{Acknowledgements}

\paragraph{}
I want to thank to my supervisor, Associate Prof. Dr. Marius Leordeanu, who brought his contribution to this master thesis and who helped me with his advice and knowledge throughout these years.
He also taught me what doing research means and inspired me passion for computer vision.

\paragraph{}
This work was supported in part by CNCS-UEFICSDI, under project PNII PCE-2012-4-0581.

\chapter*{Abstract}

In the context of the highly increasing number of features that are available nowadays we design a robust and fast method for feature selection. The method tries to select the most representative features that are independent from each other, 
but are strong together. We propose an algorithm that requires very limited labeled data (as few as one labeled frame per class) and can accommodate as many unlabeled samples. We also present here the supervised approach from which we started.
We compare our two formulations with established methods like AdaBoost, SVM, Lasso, Elastic Net and FoBa and show that our method is much faster and it has constant training time. Moreover, the unsupervised approach outperforms all the methods with which we
compared and the difference might be quite prominent. The supervised approach is in most cases better than the other methods, especially when the number of training shots is very limited. All that the algorithm needs is to choose from a pool of
positively correlated features.
The methods are evaluated on the Youtube-Objects dataset of videos and on MNIST digits dataset, while at training time we also used features obtained on
CIFAR10 dataset and others pre-trained on ImageNet dataset. Thereby, we also proved that transfer learning is useful, even though the datasets differ very much: from low-resolution centered images from 10 classes, to high-resolution 
images  with objects from 1000 classes occurring in different regions of the images or to very difficult videos with very high intraclass variance.

\chapter{Introduction}

Nowadays people aim to enable computers and robotic systems to perform tasks at the level of human beings.
Day by day, computers are becoming more powerful in terms of computational speed, storage capabilities and software intelligence. Computers have started 
to impact virtually all aspects of human life, from helping the development of other scientific fields to improving different industrial, agricultural, 
medical and consumer sectors. Ultimately they are helping us improve the quality of our lives.

While enjoying successes on all aspects of technology, automatic computer systems are still far from being able to see at the level of the human eyes 
and brain. More than $80\%$ of the information that humans process in their brains comes from sight.
Even though we are capable to see and interpret the information we gain and learn by vision from the very first months of our lives, the process that 
takes place behind is still difficult to explain. There is no general algorithm found yet that solves this problem, and the task has not yet been formalized.

In this thesis we tackle an important task in visual learning and recognition, which is that of feature selection and classifier learning with minimal 
supervision. We provide an efficient solution to this problem, and show through extensive experiments that our approach outperforms established algorithms 
from the feature selection and classification literature, such as AdaBoost, SVM, and Lasso with different types of regularization. Our method is efficient 
and general enough to be applied to various domains and here we focus on recognition especially in video, but we also provide encouraging experiments on 
image classification.
Our approach is unique especially in the fact that it focuses on learning from limited data, while performing selection at the same time. This is 
different and complementary to the current trend in learning with Deep Neural Networks. As we will discuss in the future work chapter, we envision a 
natural combination of our approach to learning and selection from limited labeled training samples with the deep hierarchical classifier structure 
paradigms, which, as of now, require huge amounts of supervised training samples.

\begin{figure}
\begin{center}
\includegraphics[scale = 1]{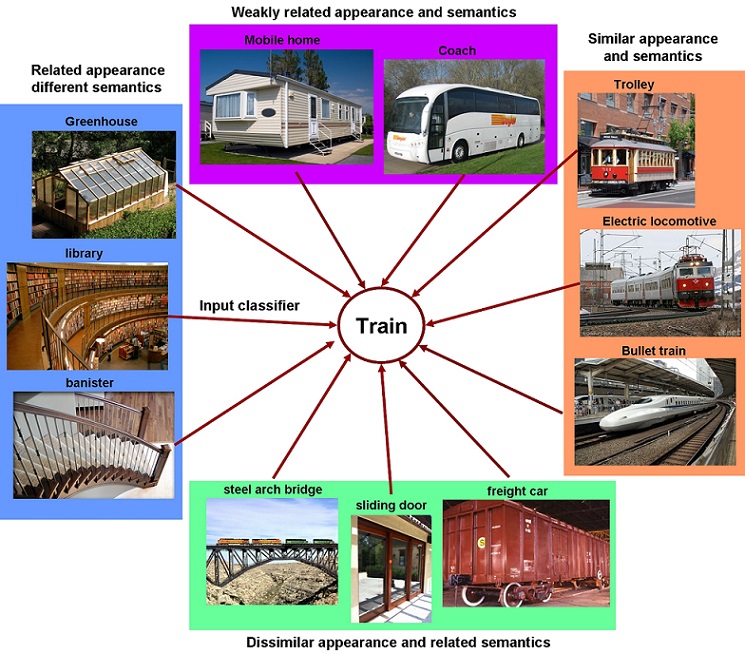}
\caption{What classes can \emph{trigger} the idea of
a ``train''? Many classes have similar appearance but are semantically less related (blue box);
others are semantically closer but visually less similar (green box). There is a continuum that relates appearance, context and semantics. Can we find a 
group of classifiers, which
are together robust to outliers, over-fitting and missing features?
Here, we show classifiers that are consistently
selected by our method from limited training data as
giving valuable input to the class ``train''.
}
\label{fig:teaser_train}
\end{center}
\end{figure}

We start from the fact that we can take advantage of the increasing number of features that can be computed on images. These features range from manually designed ones, as HOG,
SIFT and color histograms to automatically designed features generated by deep learning methods. As time complexity is still so important despite the evolution of processors and the multiprocessing capabilities available now, we cannot afford to use all these
features without a prior selection, because the task would become prohibitively costly. There is also a memory (space) problem caused by the need to store so much data. To make a clearer idea about the highly increasing number of features, we mention that
in 1997 there were few domains that used more than 40 features, while now hundreds to tens of thousands of features are explored as mentioned in~\cite{guyon2003introduction}. Other interesting facts presented and demonstrated in the same paper are: unsupervised learning is
desirable not only because in many situations the samples are not labeled, but it is less prone to overfitting, one feature that alone is useless, together with others may bring an important contribution.
Therefore, we thought about designing a method for selecting those features that
preserve as much as possible from the potential of the whole pool of features, but optimal feature selection is a NP-hard problem~\cite{guyon2003introduction,ng1998feature}. Efficient feature selection algorithms must be created so that each class 
is \emph{triggered} by a limited number of key input features (Fig.~\ref{fig:teaser_train}).

We approach feature selection from the task of discriminant linear
classification~\cite{DuHa73} with novel constraints on the solution and the features.
We put an upper bound on the solution weights and require the solution to
be an affine combination of soft-categorical features, 
which should have stronger outputs on the positive class vs. the negative. 
We term these \emph{signed features}.  Recent work~\cite{arora2013provable} 
has examined  correlations 
between the firing of units in deep neural networks for learning; 
we concentrate on much simpler relations: the sign between a feature and the target class, \emph{i.e.}
 the difference in the mean feature response over positive samples vs.\ negative ones.

We present both a supervised and an unsupervised
approach.
Our supervised method is a convex constrained 
minimization problem.
We extend this formulation to the unsupervised learning case,
with a clustering formulation that can accommodate
large amounts of unlabeled data; 
the only information required is 
the \emph{signs} of individual features, as defined shortly.
Interestingly, the latter
learning formulation is a concave minimization problem
different from the convex supervised one.
Both formulations
have important sparsity and optimality properties as well as strong generalization
capabilities in practice.
The proposed learning schemes also serve as feature  
selection mechanisms, 
such that the majority of features with zero weights can be safely ignored
while the remaining few features work together as a powerful classifier ensemble.
Consider Fig.~\ref{fig:teaser_train}: here we use image-level CNN classifiers~\cite{jia2014caffe}, pre-trained on ImageNet, to recognize trains in video frames from the YouTube-Objects dataset as done in~\cite{prest2012learning}.
Our method discovers a feature ensemble from a
pool of $6000$ classifiers that are  relevant to the concept.
Since each classifier corresponds to one ImageNet concept, we show a few of the classifiers
that have been consistently selected by our method
over $30$ trials from different small training sets of
$8$ video shots per class.

Ensemble learning proved to be a very effective method because it unites the capabilities of
simple classifiers in a more complex and powerful one. To have many images is easy, but to have many labeled images is a more expensive scenario, because we need humans to label them. We try to overcome this problem by creating
an almost unsupervised algorithm for image classification. More precisely, this algorithm requires only a very limited set of labeled data and as many unlabeled samples. We arrived to the idea that it would be much easier to work with
\emph{labeled features} than with \emph{labeled data} because the number of features is usually much smaller than the number of examples. In this thesis we will refer by \emph{labeled/signed/flipped} to those features that are positively correlated 
with the examples. Positively correlated means that the mean on the positive samples of the given feature is higher than the mean on the negative samples. We force the features to be positively correlated by changing the sign of the features that
do not comply with this condition. We start from the simple least squares formulation, then we add some constraints that ensure the sparsity of our solution and then apply the integer projected fixed point method for optimization. All that our
algorithm needs is the pool of positively correlated features and then it can use only unsupervised learning. The method is described in great detail in Chapter~\ref{cpt:pb_desc}.

Despite the small quantity of labeled data, as few as one frame per class, our algorithm manages to surpass many well-known methods like SVM, AdaBoost, Lasso etc especially in the case of a limited number of
labeled data. In Chapter~\ref{cpt:experiments} we present a series of various experiments executed.
We prove that our method is more accurate than many other methods, the training time is smaller and we also analyze other aspects of the method. The difference in performance between our method and the others is more visible in the case of very 
limited training data.

The main contributions of our approach are: 
\begin{enumerate}
 \item An efficient method for joint linear classifier learning and feature selection. We show that, both in theory and practice, our solutions are sparse. 
The number of features selected can be set to $k$ and the non-zero weights are equal to $1/k$. The simple solution enables good generalization and learning
in an almost unsupervised setting, with minimal supervision.
This is very different from classical regularized approaches such as the Lasso.
 \item Our formulation requires minimal supervision: namely only the \emph{signs} of features with respect to the target class, as defined in Sec.~\ref{sec:formulation}. 
These signs can be estimated from a small set of labeled samples, and once determined,
our method can handle large quantities of unlabeled data with excellent accuracy and generalization in practice.
 \item In our extensive experiments, both supervised and unsupervised variants of our method demonstrate superior performance in terms of learning time and accuracy when compared to established approaches such as AdaBoost, the Lasso, Elastic Net and SVM. In particular, 
the unsupervised variant significantly outperforms its competitors and is the least sensitive to the quantity of labeled data.
\end{enumerate}

The outline of this work is as follows. In Chapter~\ref{cpt:relWork} we present other existent approaches to our problem, in Chapter~\ref{cpt:pb_desc} we explain our own approach for both supervised and unsupervised variants and make a theoretical 
analysis of the algorithms proposed. In Chapter~\ref{cpt:experiments} we talk about the features used and the experiment settings and present a number of experiments done with our two approaches and with the methods that we compare with. 
Chapter~\ref{cpt:conclusion} concludes our work with a summary and some ideas about the research we did.  An initial version of this work appeared in our paper~\cite{leordeanu2014features}, which presents the supervised learning scheme. 
Later we developed the \emph{almost} unsupervised variant of our method, which will appear at the AAAI-16 International Conference on Artificial 
Intelligence~\cite{feature_selection_our_aaai2016} along with the initial supervised case. An arxiv version of this paper ~\cite{aaai2016arxiv} also appeared. Our work also draws inspiration from ideas initially proposed in the 
Classifier Graph paper~\cite{leordeanu2014thoughts} and validates some of those concepts. For example, in our experiments we demonstrate that it is possible to 
learn powerful novel classifiers by efficiently selecting and combining previously learned independent classifiers from a very large heterogeneous pool, which is 
one of the ideas presented in~\cite{leordeanu2014thoughts}.

\chapter{Related work}
\label{cpt:relWork}
In this chapter we are going to make a presentation of the main approaches to the feature selection problem, and we will focus on its application in the domain of computer vision, more precisely object classification. We will talk about regularized
approaches to feature selection like: Lasso and Elastic Net, to ensemble learning methods like: bagging, boosting and decision trees ensembles. We will not limit to feature selection, we will also refer to some of the descriptors used in computer vision:
HOG and SIFT. We used the former in our implementation. We will also talk about object recognition, more precisely about part based models.

%\section{Related work to feature selection}
\paragraph{\textbf{Relation to feature selection through regularization:}}
Feature selection is such a general topic that it is used in many domains, from machine learning and bioinformatics to data mining and computer vision. Feature selection is a procedure used in machine learning for extracting a subset of relevant features in order
to create a less complex model of learning. The motivation behind this technique is the fact that when the initial pool consists of a huge number of features, some of them are redundant (they do not bring any useful information). When we have 
a huge number of features and a small number of examples, we need to discard some of the features in order to avoid overfitting. Being such a vast topic, there are many approaches to feature selection: from ensemble learning methods, to 
$L_1$, $L_2$ regularization, greedy selection~\cite{couvreur2000optimality}, genetic algorithms~\cite{guo2011genetic} (optimize feature selection in accordance with the resource limitations) and ant colony optimization~\cite{kabir2012new}. Some benefits of feature selection techniques are mentioned in~\cite{chandrashekar2014survey}: better understating 
of the data, curse of dimensionality reduction, improvement of predictor performance as noisy features can degrade the learning process. According to~\cite{chandrashekar2014survey} the feature selection methods can be divided into two classes: filter and wrapper. The former refers to those methods that
evaluate each feature individually, rank the features and give up those that are below a threshold. The later category refers to evaluating subsets of features and deciding which one is the best. Because the number of subsets is $2^N$, an
exhaustive search is impossible, thus some search algorithms are used to maximize the objective function.

$L_0$ regularization is more suitable for feature selection than any other regularization type because $L_0$ norm is equal to the number of non-zero values. In feature selection we should minimize the number of features selected, namely the $L_0$-norm.
As stated in~\cite{le2013efficient, liu2014efficient}, $L_0$ regularization leads to very difficult minimization problems, therefore it is common to use other types of regularization as $L_1$ or $L_2$. In fact, $L_0$ optimization is an NP-hard problem. $L_1$
regularization is known to be the best relaxation of $L_0$ regularization, being solved by gradient descent.

$L_1$ and $L_2$ regularization are used for feature selection in~\cite{xu2014gradient, zhang2013multi, nie2010efficient}. $L_1$ regularization, usually referred as Lasso~\cite{tibshirani1996regression}, has three main benefits: it avoids overfitting by 
penalizing big weights, leads to sparse solutions and is a convex optimization problem, which has some disadvantages. If the number of features $p$ is greater than the number of samples $n$, then Lasso will select at most $n$ features and it
fails to select grouped features - if a group of features are related, then it selects only one of them. Another regularized approach  named ridge regression uses the $L_2$-norm. Its drawbacks come from the
fact that it only shrinks some parameters towards zero without setting them to exactly 0, thus all the features are kept, none of them being discarded. This makes the model difficult to interpret. Elastic Net~\cite{zou2005regularization} combines these two methods and benefits 
from their advantages. The Elastic Net regularization is expressed by: \mbox{$\beta^* = \argmin_\beta\|y-\mathbf{X}\beta\|^2+\lambda_2\|\beta\|^2+\lambda_1\|\beta\|_1$}. The $L_1$ regularization term enforces the sparsity of the solution, while $L_2$ 
regularization term removes the limitation on the number of selected features and encourages grouping effect.

%$\|\mathbf{Fw} - \mathbf{t}\|^2=\mathbf{w^\top (F^\top F) w} - 2\mathbf{(F^\top t)}^\top \mathbf{w} + \mathbf{t}^\top \mathbf{t}$

In~\cite{xu2014gradient}
the authors use decision trees for feature selection with $L_1$ regularization. Each tree uses some features in its construction and the total weight for each feature is given by the sum of the weights of the threes in which that feature appears. The optimization problem 
is non-convex and non-differentiable, but up to a fixed point it can still be optimized with gradient boosting. At each iteration only one dimension of the weights is optimized. In~\cite{zhang2013multi}, the authors try to recover $w$ given their input vectors
and the output variables. They assumed that the weight matrix is sparse, and they focus on estimating the non-zero weights. The $L_1$ regularized formulation of the problem can be solved with convex programming. It was observed that $L_1$ regularization
leads to sparse solutions. In~\cite{le2013efficient}, $L_2$-$L_0$-norm  is used in the optimization problem and the authors try to solve it by DC programming (Difference of Convex functions) and approximate the $L_0$-norm by a concave function. They want to select features
for multiclass SVM. The objective function is decomposed as a difference of convex functions and the solution is obtained iteratively.

\paragraph{\textbf{Relation to ensemble learning:}}
During the last decades of research in machine learning, it has been observed that a combination of multiple individual classifiers is stronger than
a single classifier. As stated in~\cite{dietterich2000ensemble}, the two necessary conditions for an ensemble to work are the following: 1) the base classifiers must be better than random and 2) the base classifiers must be as diverse as possible. If the first condition does not hold, then
the ensemble will have a higher error than the base classifiers. If the second condition is not true and the classifiers are identical (or very similar), they will make the same errors and the ensemble will do the same. The result of the ensemble
classifier is the outcome of a voting procedure of the composing classifiers. If all vote the same, then the ensemble will bring no improvement.
The main approaches to ensemble learning~\cite{maclin2011popular, buhlmann2012bagging} are  boosting~\cite{freund1997decision}, bagging 
\cite{breiman1996bagging} and decision trees ensembles~\cite{breiman2001random, kwok2013multiple}. Selecting a subset of features from the whole pool presents two main advantages over the approach in which all the features are considered as
stated in~\cite{yang2011l2}. First, the computational costs are much more reduced. Second, the noisy features are removed and the algorithm can generalize better when it has to deal only with more representative features.

Bagging was the first efficient machine learning ensemble method. It is a meta-algorithm whose main idea is to train more classifiers on different 
subsets of the training set. The classifiers are trained independently, even in parallel. The subsets are obtained by sampling with replacement examples from the whole set. For each subset created, a new classifier is trained
and finally the results of the classifiers are averaged in order to obtain a single result (regression) or they are subject to a voting procedure
(classification). The advantage of this method are the ability to reduce variation and to improve performance for unstable classifiers that vary 
significantly when small changes in the data appear. A particularity of bagging that is different in the case of our method is the fact that the former averages the whole set of classifiers, while
the later uses only a subset of them. We believe that fewer classifiers could be better than all, this idea being also supported in~\cite{zhou2002ensembling}, where an algorithm GASEN-b is proposed. In this algorithm a weight is assigned to
each classifier. The weights are optimized using genetic algorithms and then, the classifiers whose weights were greater than a threshold are selected.

Boosting is a machine learning ensemble meta-algorithm. Through boosting a set of weak classifiers are combined in order to form a strong one. It 
is a sequential algorithm in which at each iteration the weak classifiers that have not been chosen yet are trained and the one that performs
best in the given state is chosen. The goal is to choose the classifiers so as to avoid redundancy as much as possible. The examples in the training set are weighted, so that those that are more difficult will weight more, while
the simpler ones will weight less; those examples that were misclassified at a given iteration will weight more in the next ones, so that the new classifiers focus more on them. There are many kinds of boosting algorithms: AdaBoost, 
LPBoost, LogitBoost etc. A weakness of the method is the
fact that it performs well only when the base classifiers are weak, it cannot take advantage of stronger base classifiers. This is different in 
the case of our method, because it can work very well also with stronger classifiers. Other problems are the fact that initial classifiers tend to have a much greater importance
than those that are chosen in later stages of the algorithm and the training phase is very slow. This method became famous through its particularization AdaBoost when it was successfully used for face detection by Viola and Jones 
in~\cite{viola2004robust}.

Decision trees ensembles represent a machine learning method that constructs a multitude of decision trees at training time and the result of 
the method is the average of the results of these trees (regression) or the mode of their results (classification). This method solves the problem
of individual decision trees of overfitting. Unlike decision trees ensembles, our method averages only a subset of classifiers, not the whole pool,
which makes a big difference.

\paragraph{\textbf{Image descriptors:}}
%\section{Related work to image descriptors}
As stated before, feature selection is a very general problem applied in many domains. As in computer vision we do not usually work directly with the raw pixels, we should apply some descriptors on the image and then use their values as features.
Given that the number of descriptors has increased very much during the last years, we should select only a part of them in order to apply
then some machine learning algorithms. Image descriptors can be automatically or manually designed. Automatically designed image descriptors can be obtained using deep neural networks, such a network is described in~\cite{jia2014caffe}. Such
descriptors are very numerous, for example with Caffe we can obtain 1000 descriptors. On the other hand, there are fewer manually designed descriptors, but stronger. We can mention Scale Invariant Feature Transform (SIFT)~\cite{lowe2004distinctive} - more suitable for 
object recognition, Histograms of Oriented Gradients (HOG)~\cite{dalal2005histograms} - better for object classification and color histogram descriptors as: Dominant Color, CSD, CLD, all presented in~\cite{min2009effective}. In the following paragraphs we present the most important
among them.

David Lowe proposes in~\cite{lowe2004distinctive} a new method for object recognition that
consists in identifying a set of features in images that are invariant to scaling, ro-
tation, translation, illumination, some affine transforms and also are sufficiently
distinctive. The main difficulty in object recognition is the identification of such
features. In each image a number of the order of 1000 of such features are identi-
fied in less than a second. The nearest-neighbour algorithm is used to identify the
model that best matches the current object. If the model and the object have at
least 3 features in common, then the object is considered recognized. Given that
the number of initial features for each object is about 20, this approach is effective
also when the object is partially occluded because of the fact that only three key
points are necessary to identify the object. The most frequent approaches at that
time were based on template matching, but this solution was good only for the
situations in which the illumination and the position of the object were invariant.
Another solution was to identify only the features that are invariant to certain
conditions and template match only on these: line segments, regions, grouping
of edges. Another kind of approach was to detect the corners in an image and to
focus the matching only on these areas, not on the whole image. The weakness
of the last approach stands in the fact that it is not invariant to scaling, this means
that the image has to be scanned at many scales in order to obtain a reliable
result. In order to ensure scale and rotation invariance, an image pyramid is
built to create the scale space. The scale space of the image is built and the
difference-of-gaussian of the images obtained is computed. For each point, a 3
by 3 vicinity is considered from 3 neighbouring scales. If a point is a minima or
a maxima in the three considered images, then it is an interest point. After the
interest points are identified, the nearest neighbour algorithm is used to group the
SIFT features and to detect the objects found in the image. Each key feature in
the image has associated an orientation and based on these orientations, a his-
togram of orientations is built using 36 bins that cover the range of 360 degrees.
Lowe's approach has similarities with the mechanisms that primates use in order
to recognize an object. Neuroscientists discovered neurons that are specialized
in recognizing shapes like dark five-sided star shape, or a circle with a protrud-
ing element. These being intermediate features that help primates together with
color and texture cues to recognize objects.

Dalal and Triggs in~\cite{dalal2005histograms} have designed an algorithm for pedestrian 
detection. Despite the fact that they illustrated this method on pedestrians, it is
a general object category detection algorithm that can be used very well for any category of objects. 
The idea of the authors was to find a representation that helps
discriminating the human body over the cluttered background. HOG used to
perform better than the existing methods. The authors of the study analysed
every aspect of the algorithm in order to obtain the best possible results: the
number of bins, the number of horizontal/vertical cells and the local normalization. 
They use a histogram of gradients in order to create an encoding of the
image. For each training image the histogram of oriented gradients is computed.
When an image should be tested, the same histogram is created and the image is
classified, based on its histogram as containing a pedestrian, or not. The gradients 
of the image are computed, then the image is divided in equal cells. In each
cell, each pixel will vote for a certain orientation of the edges. 
are cumulated in orientations bins. Figure 3.1 shows such a histogram computed
for a pedestrian.Usually the range is between 0 and 180 degrees and the number
of bins chosen by the authors was 9. It is a weighted vote in which each pixel
contributes with the magnitude value. A local normalization step follows, when
the cells are grouped in blocks, and they are normalized with regard to each
block in which they appear. The blocks can or cannot overlap, in the former case
the histogram will be bigger because some cells will be repeated, but the results
proved to be better. The optimum parameters observed were: blocks of 3 x 3
cells, 9 bins and 6 x 6 pixels cells. They used linear SVM for classification. In
order to detect pedestrians in each position and for each scale, Dalal and Triggs
used a sliding window approach in which a filter is applied for all positions and
for all scales. The results were very good, on the MIT database the accuracy was
almost 100\%, therefore they created a more challenging dataset with people in
different position and different backgrounds. On INRIA set it obtains FPPW(false positives per window) score an order
of magnitude better. HOG outperformed the other algorithms: wavelet,
PCA-SIFT and Shape Context. In this paper, Dalal and Triggs showed that histograms of oriented gradients,
locally contrast normalized can give very good results compared to the methods
that existed at that moment. They also showed that the simplest methods of
computing the gradients are the most performant, but the performance is highly
influenced also by the binning, the contrast normalization and by the overlapping
of the blocks.

%un descriptor de culoare
\paragraph{\textbf{Relation to object recognition:}}
During the last years, image classification evolved towards more complex higher-level models.
Such an example is a method designed by Felzenszwalb et al. that learns \textit{part-based models} and that we present in the following lines.
In their study~\cite{felzenszwalb2010object}, Felzenszwalb et al. try to address the problem of the
gap between the performance obtained using part-based models and other much
simpler methods that outperform it on difficult datasets. This problem is caused
by the fact that an object can be viewed from different orientations and thus,
they propose to use mixture models to deal with this kind of situations. Complex 
part-based models being outperformed by simpler methods on more difficult datasets is due to the
fact that they are more difficult to train, because learning methods like SVM cannot be used. A
star model of the object is built using a coarse filter as root and finer filters to
detect parts of the object. The model is represented as a tuple of the root part
and the smaller parts. Each part is composed from its filter, the anchor (where
it should be placed relative to the main part) and the penalties for changes in
position. The score of an object is computed as the sum of the scores of each
filter minus the deformation cost of each part relative to its expected position.
The root location with highest scores represent detections and together with the
placement of the parts constitute an object hypothesis. The system was tested
on Pascal VOC dataset and also on INRIA, on 20 categories of objects.

\chapter{Problem description}
\label{cpt:pb_desc}

In this chapter we will present our approach to feature selection, the mathematical formulation in which we start from the least squares formulation and add new constraints that lead to important theoretical properties. We also talk about
our two methods: the supervised and the unsupervised one which is derived from the former. We present our algorithms and discuss the intuition behind our approach. We also make a theoretical analysis of the method.

\section{Our method}
\label{sec:formulation}

Through our method we approach the case of binary classification, while for the multi-class scenario we apply the one vs.\ all
strategy.
Our training set is composed of $N$ samples, each $i$-th sample being expressed
as a column vector $\mathbf{f}_i$ of $n$ features with values in $[0,1]$;
such features could themselves be outputs of classifiers.
We want to find a 
vector $\mathbf{w}$, with elements in $[0,1/k]$ and unit $L^1$-norm, 
such that $\mathbf{w}^T\mathbf{f}_i \approx \mu_P$ when the $i$-th sample is
from the positive class and $\mathbf{w}^T\mathbf{f}_i \approx \mu_N$ otherwise, with $0 \leq \mu_N < \mu_P \leq 1$.
For a positive labeled training sample $i$, we 
fix the ground truth target $t_i = \mu_P = 1$ 
and for a negative one we fix it to $t_i = \mu_N = 0$. 
Our novel constraints on $\mathbf{w}$ 
limit the impact of each individual feature $f_j$, encouraging the selection of features that are powerful in combination, with no single one strongly dominating. This produces solutions with good generalization power.
In Sec.~\ref{sec:algorithms} we
show that $k$ is equal to the number of selected features, all with 
weights $=1/k$. The solution we look for is 
a weighted feature average 
with an ensemble response that is stronger on positives than on negatives.
For that, we want any feature $f_j$ to have expected value $E_P(f_j)$ over positive samples 
greater than its expected value $E_N(f_j)$ over negatives.
From the labeled samples we estimate the sign of each feature $\mathit{sign}(f_j) = E_P(f_j) - E_N(f_j)$
and if it is negative we simply \emph{flip} the feature values: $f_j \gets 1 - f_j$.  

\subsection{Supervised learning}
We begin with the supervised learning task, which
we formulate as a least-squares constrained minimization problem.
Given the $N \times n$ feature matrix $\mathbf{F}$ with 
$\mathbf{f}^\top_i$ on its $i$-th row and the ground truth vector $\mathbf{t}$, we look for
$\mathbf{w^*}$ that minimizes
$\|\mathbf{Fw} - \mathbf{t}\|^2=\mathbf{w^\top (F^\top F) w} - 2\mathbf{(F^\top t)}^\top \mathbf{w} + \mathbf{t}^\top \mathbf{t}$, and obeys the required constraints. We drop
the last constant term $\mathbf{t}^\top \mathbf{t}$
and obtain the following convex minimization problem:
\begin{eqnarray}
\label{eq:learning}
\mathbf{w^*} & = & \argmin_w J(\mathbf{w}) \\
    & = & \argmin_w \mathbf{w^\top (F^\top F) w} - 2\mathbf{(F^\top t)}^\top \mathbf{w} \nonumber \\
    & & s.t. \sum_j w_j =1 \;, \; w_j \in [0,1/k]. \nonumber
\end{eqnarray}
\noindent
Our least squares formulation is related to Lasso, Elastic Net and other regularized approaches, with the  distinction that in our case
individual elements of $\mathbf{w}$ are restricted to $[0,1/k]$, which
leads to important theoretical properties regarding sparsity and directly impacts generalization power (Sec.~\ref{sec:algorithms}). This 
also leads to our (almost) unsupervised approach, presented in the next section.

%As we will see, in this way we have full control over the number
%of feature selected. We also demonstrate better generalization performance
%and weaker sensitivity to the number of features.
%Since $\mathbf{t}$ is the ground truth, the last term is constant.
%After dropping it, we note that the supervised learning task is a special case
%of clustering with pairwise and unary terms, as defined in
%~\cite{key:bulo_nips09,key:latecki_nips10,leordeanu2012efficient}.
%Note that our formulation can be easily changed into a concave maximization
%problem by changing the signs of the terms.
%Since the algorithm of~\cite{leordeanu2012efficient} works with both positive and negative
%terms, we adapt their efficient optimization scheme that achieves near-optimal solutions
%in only $10-20$ iterations.

\subsection{Unsupervised learning}

Consider a pool of signed features correctly flipped according to their signs, which could be known a priori, or estimated from a small set of 
labeled data. We make the simplifying assumption that the signed features'
expected values for positive and negative samples, respectively,
are close to the ground truth target values
$(\mu_P, \mu_N)$. %This can
%be approximated by appropriate normalization, estimated 
%from the small supervised set. 
%This can be estimated from a small supervised set.  
Then, for a given sample $i$, and any 
$\mathbf{w}$ obeying the constraints, the expected value of the weighted average $\mathbf{w}^\top\mathbf{f}_i$ is also close to the ground truth target $t_i$:
$E(\mathbf{w}^\top\mathbf{f}_i)=\sum_j w_j E(\mathbf{f}_i(j)) 
\approx (\sum_j w_j)t_i = t_i$.
Then, for all samples we have the expectation
$E(\mathbf{F}\mathbf{w}) \approx \mathbf{t}$, such that any feasible 
solution will produce, on average, approximately correct answers. 
Thus, we can regard the supervised learning scheme as attempting to reduce the variance of the feature ensemble output, as their expected value
is close to the ground truth target. If we now introduce the approximation
$E(\mathbf{F}\mathbf{w}) \approx \mathbf{t}$ into the 
learning objective $J(\mathbf{w})$, we obtain our new ground-truth-free objective $J_u(\mathbf{w})$ with the following 
learning scheme, which is unsupervised once the feature signs are determined.
Here $\mathbf{M = F^\top F}$:
\begin{eqnarray}
\label{eq:semisup_learning}
\mathbf{w^*} & = & \argmin_w J_u(\mathbf{w} )\\
    & = & \argmin_w \mathbf{w^\top (F^\top F) w} - 2\mathbf{(F^\top (\mathbf{F}\mathbf{w}))}^\top \mathbf{w} \nonumber \\
    & = & \argmin_w (-\mathbf{w^\top (F^\top F w})) = \argmax_w \mathbf{w^\top M w} \nonumber \\
    & & \text{s.t.} \sum_j w_j =1 \;, \; w_j \in [0,1/k]. \nonumber
\end{eqnarray}
Interestingly, while the supervised case is a convex minimization problem,
the unsupervised learning scheme
is a concave minimization problem, which is NP-hard. 
This is due to the change in sign of the matrix $\mathbf{M}$. 
However, since $\mathbf{M}$ in the unsupervised case could be
created from larger quantities of unlabeled data, $J_u(\mathbf{w})$ could
in fact be less noisy than $J(\mathbf{w})$ and produce significantly
better local optimal solutions --- a fact confirmed by our experiments.

\subsection{Intuition}

Let us take a closer look at the two terms involved in our objectives, 
the quadratic term: $\mathbf{w^\top M w}=\mathbf{w^\top (F^\top F) w}$ 
and the linear term: $\mathbf{(F^\top t)}^\top \mathbf{w}$. 
If we assume that feature outputs have similar expected values, 
then minimizing the linear term in the supervised case 
will give more weight to features that are strongly correlated with the ground truth and are good
for classification, even independently. 
However, things become more interesting when looking at the role played by the quadratic term in the two cases of learning.
The positive definite 
matrix $\mathbf{F^\top F}$ contains the dot-products between pairs of feature responses over the samples.
In the supervised case, minimizing $\mathbf{w^\top (F^\top F) w}$
should find \emph{a group of features}
that are as uncorrelated as possible. Thus we seek group of features
that are individually relevant due to the linear term, 
but not redundant with respect to each other due to the quadratic term. 
They should be \emph{conditionally independent} given the class, an observation that is consistent with earlier
research in machine learning (e.g.,~\cite{dietterich2000ensemble})
and neuroscience (e.g.,~\cite{rolls2010noisy}).
% The idea is also related to the recent work on discovering discriminative groups of HOG filters~\cite{ahmed2014knowing}.

In the unsupervised case, the task seems reversed: maximize the same
quadratic term $\mathbf{w^\top M w}$, with no linear term involved. We could interpret this as 
transforming the learning problem into a special case of clustering with pairwise constraints, related to methods such as spectral clustering with
$L^2$-norm constraints~\cite{key:sarkar} and robust hypergraph clustering with 
$L^1$-norm constraints~\cite{key:bulo_nips09,key:latecki_nips10}. 
The problem is addressed by finding the group of features with strongest 
intra-cluster score --- the largest amount of covariance. In the absence of
ground truth labels, if we assume that features in the pool are, 
in general, correctly signed 
and not redundant, 
then the maximum covariance is attained by features
whose collective average varies the most as the hidden class labels also vary.
Thus, the unsupervised variant seeks features that respond in a united manner
% that is sensitive to the natural variation in
to the distributions of the two classes.

% In other words, the variation in the good features caused by 
% the natural differences between the two classes
% is the strongest when compared to noisy
% causes of variation that are more or less independent.

\section{Algorithms}
\label{sec:algorithms}

In our both approaches, we first need
to determine the \emph{sign} for each feature, as defined before.
Once it is estimated, we can set up the optimization
problems to find $\mathbf{w}$. In Algorithms
~\ref{alg:sup_learning} and ~\ref{alg:semisup_learning}, we present our supervised and unsupervised learning methods. 
The supervised case is a convex minimization problem, with efficient global optimization possible in polynomial time. 
The unsupervised learning is a concave minimization problem, which is NP-hard and can only have local efficient optimization.

\begin{algorithm}
%\caption{Weakly supervised learning}
\caption{Supervised learning}
\label{alg:sup_learning}
\begin{algorithmic}
\STATE Learn feature signs from the labeled samples.
\STATE Create $\mathbf{F}$ with flipped features from the labeled samples.
\STATE Set $\mathbf{M} \gets \mathbf{F^\top F}$, \\
\STATE Find  $\mathbf{w^*} = \argmin_w \mathbf{w^\top M w} - 2\mathbf{(F^\top t)}^\top \mathbf{w} \nonumber$ \\
       $\; \; \; \; \; \; $ s.t. $\sum_j w_j =1 \;, \; w_j \in [0,1/k]$.
\RETURN $\mathbf{w^*}$
\end{algorithmic}
\end{algorithm}

\begin{algorithm}
%\caption{Weakly supervised learning}
\caption{Unsupervised learning from signed features}
\label{alg:semisup_learning}
\begin{algorithmic}
\STATE Learn feature signs from a small set of labeled samples.
\STATE Create $\mathbf{F}$ with flipped features from unlabeled data.
\STATE Set $\mathbf{M} \gets \mathbf{F^\top F}$, \\
\STATE Find $\mathbf{w^*} = \argmax_w \mathbf{w^\top M w}$ \\ 
       $\; \; \; \; \; \; $ s.t. $\sum_j w_j =1 \;, \; w_j \in [0,1/k]$.
\RETURN $\mathbf{w^*}$
\end{algorithmic}
\end{algorithm}

%Here note a very interesting connection with a possible spectral
%feature selection algorithm (here the work of Zhao should be cited,
%even though it is different .... he even has a book on Spectral Feature Selection ...)
%, where under the L2-norm constraint
%the optimum $\mathbf{w^*}$ is the principal eigenvector of $\mathbf{M}$.

\begin{figure}
\begin{center}
\includegraphics[scale = 0.6, angle = 0, viewport = 0 0 500 520, clip]{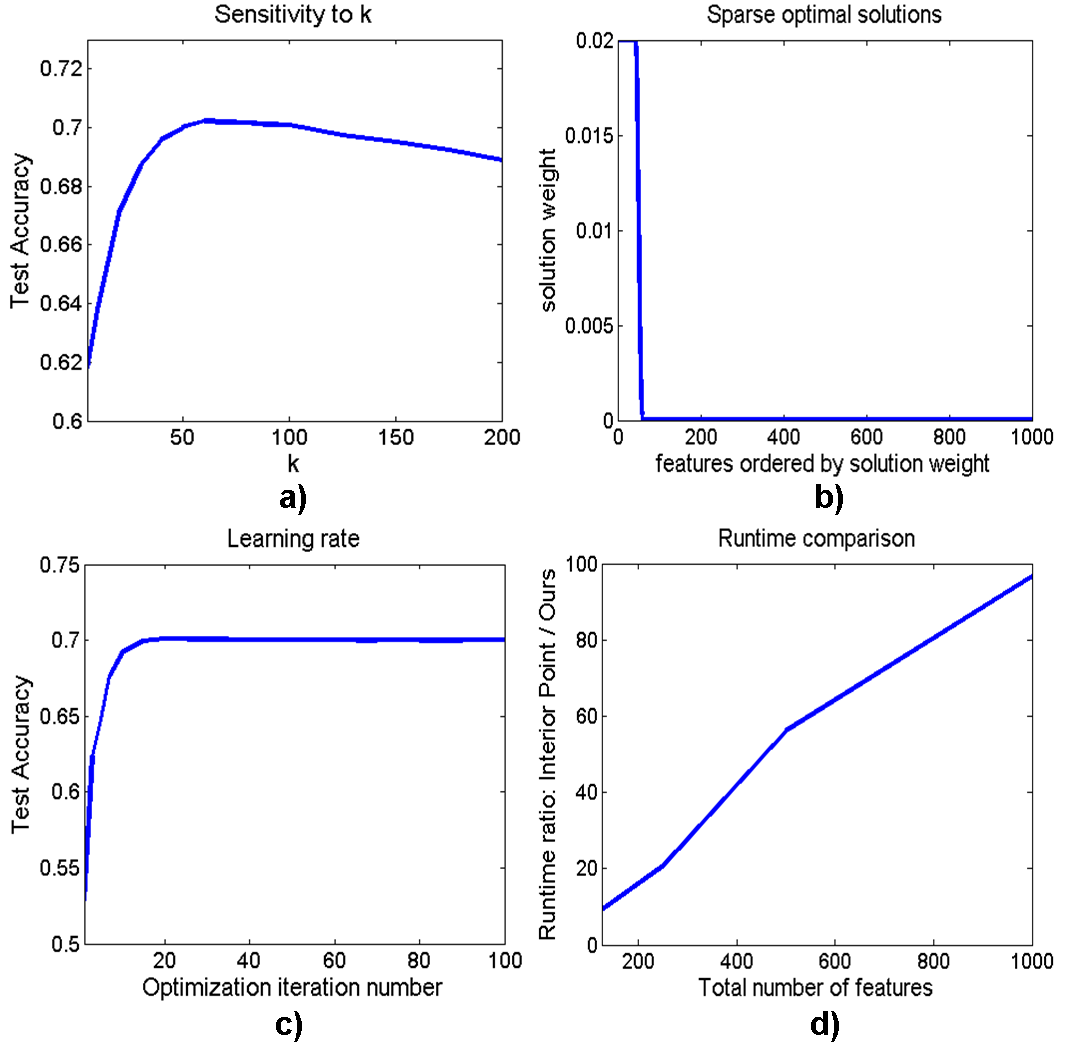}
\caption{Optimization and sensitivity analysis:
a) Sensitivity to $k$.
Performance improves as features are added, is stable
around the peak $k=60$ and falls for $k>100$ as useful features
are exhausted.
b) Features ordered by weight for $k=50$ confirming that our method selects
equal weights up to the chosen $k$.
c) Our method almost converges in $10$--$20$ iterations.
d) Runtime of interior point method divided by ours,
both in Matlab and with $100$ max iterations. All results are averages
over $100$ random runs.
}
\label{fig:optimization}
\end{center}
\end{figure}

There are many possible fast methods for optimization. 
In our implementation we adapted the integer projected
fixed point (IPFP) approach~\cite{leordeanu2012efficient,key:leordeanu_IPFP},
related to the Frank-Wolfe algorithm\cite{key:FW},
which is efficient in practice (Fig.~\ref{fig:optimization}c)
and is applicable to both supervised and unsupervised cases.
The method converges to a stationary point --- the
global optimum in the supervised case.
At each iteration IPFP approximates
the original objective
with a linear, first-order Taylor approximation that can be optimized immediately in the feasible domain.
That step is followed by a \emph{line search} with rapid closed-form solution, and the process is repeated
until convergence.
%(See~\cite{leordeanu2012efficient,key:leordeanu_IPFP} for more details).
In practice, $10$--$20$
iterations bring us close to the stationary point; 
nonetheless, for thoroughness, we use $100$ iterations
in all our experiments. See, for example,
comparisons to Matlab's \emph{quadprog} run-time
for the convex supervised learning case in Fig.~\ref{fig:optimization}
and to other learning methods in Fig.~\ref{fig:test_and_time}.
Note that once the linear and quadratic terms are set up, the learning
problems are independent of the number of samples and only dependent
on the number $n$ of features considered, since $\mathbf{M}$ is $n \times n$
and $\mathbf{F^\top t}$ is $n \times 1$.

%The computational cost of the optimization method we use is
%$O(Sn^2)$~\cite{leordeanu2012efficient},
%where $S$ is the number of iterations and $n$ is the number of features.
%In our experiments we use $S=100$, even though $S=20$ would suffice.
%The more general interior point method for convex optimization using Matlab's
%$quadprog$ is polynomial, but considerably slower than ours, by a factor that increases
%linearly with the number of features (see Fig.~\ref{fig:optimization}).
%For $125$ features it is $9$ times slower, and for $1000$ features, about $100$ times slower.

\begin{figure}
\begin{center}
\includegraphics[scale = 0.5, angle = 0, viewport = 0 0 700 280, clip]{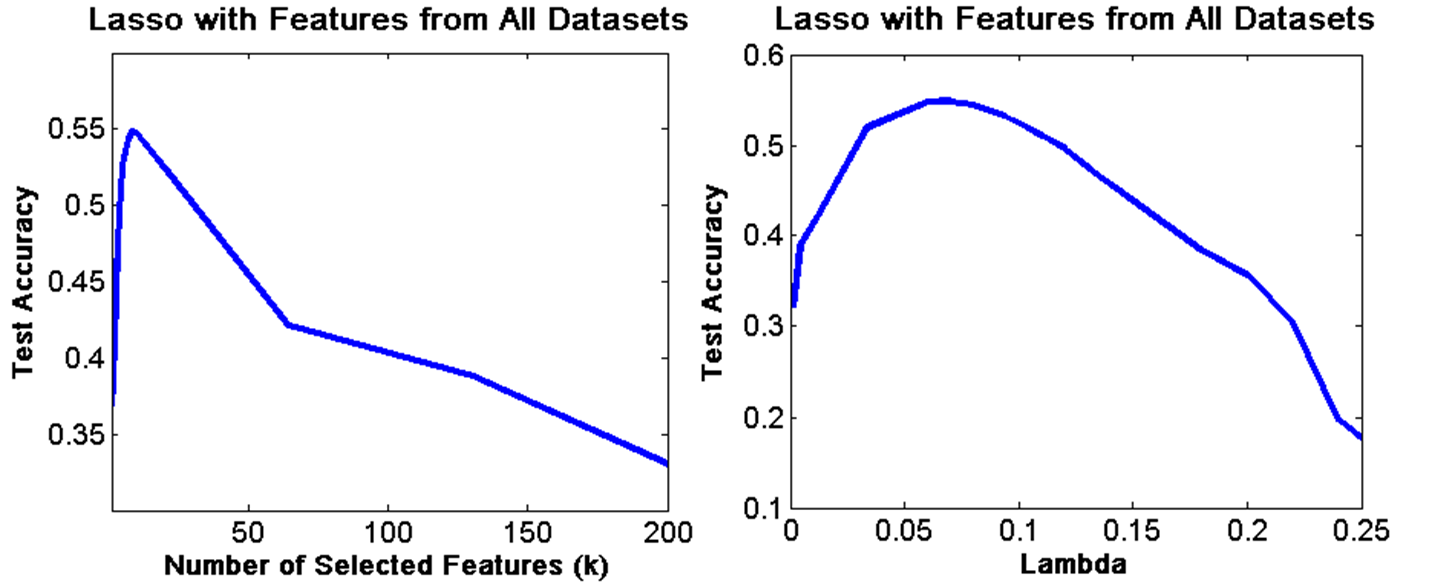}
\caption{Sensitivity analysis for Lasso:
Left: sensitivity to number of features with non-zero weights
in the solution.
Note the higher sensitivity when compared to our approach.
Lasso's best performance is achieved
for fewer features, but the accuracy is generally worse than in our case.
Right: sensitivity to lambda $\lambda$, which controls the 
L1-regularization penalty.
}
\label{fig:Lasso_sensitivity}
\end{center}
\end{figure}

\section{Theoretical analysis}
%\label{sec:theory}

First we show that the solutions are sparse
with equal non-zero weights (P1), also
observed in practice (Fig.~\ref{fig:optimization}b).
This property makes our classifier learning also
an excellent feature selection mechanism.
Next, we show that simple equal weight solutions are likely to minimize the output variance over samples of a given class
(P2) and minimize the error rate. This explains the 
good generalization power of our method. 
Then we show how the error rate is expected
to go towards zero when the number of considered 
non-redundant features increases (P3),
which explains why a large diverse pool of features is beneficial.
Let $J(\mathbf{w})$ be our objective for either the
supervised or unsupervised case:

\noindent \textbf{Proposition 1:}
Let $\mathbf{d(w)}$ be the gradient
of $J(\mathbf{w})$. The partial derivatives $d(\mathbf{w})_i$ corresponding
to those elements $w^*_i$ of the stationary points  
with non-sparse, real values in $(0,1/k)$ must be 
equal to each other.

\noindent \textbf{Proof:}
The stationary points for the Lagrangian
satisfy the Karush-Kuhn-Tucker (KKT) necessary optimality conditions.
The Lagrangian is $L(\mathbf{w}, \lambda, \mu , \beta) = J(\mathbf{w}) - \lambda (\sum w_i - 1) + \sum \mu_i w_i + \sum \beta_i (1/k - w_i)$.
From the KKT conditions at a point $\mathbf{w^*}$ we have:
\[
\begin{array}{l}
\mathbf{d(w^*)} - \lambda + \mu_i - \beta_i = 0,\\
\sum_{i=1}^{n} \mu_i w^*_i = 0,\\
\sum_{i=1}^{n} \beta_i (1/k - w^*_i) = 0.\\
\end{array}
\]
Here $\mathbf{w^*}$ and the Lagrange multipliers have non-negative elements,
so if $w_i > 0 \Rightarrow \mu_i = 0$
and $w_i < 1/k \Rightarrow \beta_i = 0$. Then there must exist
a constant $\lambda$ such that:
\[d(\mathbf{w^*})_i =  \left\{
\begin{array}{ll}
 \leq \lambda, & w^*_i = 0, \\
 = \lambda,    & w^*_i \in (0, 1/k), \\
\geq \lambda, & w^*_i = 1/k. \\
\end{array}\right.
\]
This implies that all $w^*_i$ that are different from $0$ or $1/k$
correspond to partial derivatives $d(\mathbf{w})_i$ that are equal
to some constant $\lambda$, therefore 
those $d(\mathbf{w})_i$ must be equal to each other, 
which concludes our proof.

From Proposition $1$ it follows that in the general case, when the
partial derivatives of the objective error function
at the Lagrangian stationary point are unique, the elements of
the solution $\mathbf{w^*}$ are either $0$ or $1/k$.
Since $\sum_j w^*_j = 1$ it follows that the number of nonzero weights
is exactly $k$, in the general case.
Thus, our solution is not just a simple linear
separator (hyperplane), but also a sparse representation and a feature selection
procedure that effectively averages the selected $k$ (or close to $k$) features.
The method is robust to the choice of $k$ (Fig.~\ref{fig:optimization}.a)
and seems to be less sensitive to the number of features selected 
than the Lasso (see Fig.~\ref{fig:Lasso_sensitivity}).
In terms of memory cost, 
compared to the solution
with real weights for all features, whose storage requires 
$32n$ bits in floating point representation, our averaging of $k$ selected
features needs only $k\log_2 n$ bits --- select $k$ features out of $n$ possible
and automatically set their weights to $1/k$.
% Our approach is simpler, learns faster (Fig.~\ref{fig:test_and_time}) and is memory efficient.
%They seem to follow closer the Occam's Razor principle~\cite{blumer1987occam}, which would explain in part their good generalization from limited data, with a very effective weakly supervised learning scheme.
Next, for a better statistical interpretation
we assume the somewhat idealized case when all features
have equal means $(\mu_P, \mu_N)$ and equal standard deviations
$(\sigma_P, \sigma_N)$ over positive (P) and negative (N) training sets, respectively.
%This effectively induces a bimodal distribution for the average response of $n$ features.

\noindent \textbf{Proposition 2:}
If we assume that the input soft classifiers are independent
and better than random chance, the error rate converges towards $0$
as their number $n$ goes to infinity.

\noindent \textbf{Proof:}
Given a classification threshold $\theta$ for $\mathbf{w}^T\mathbf{f}_i$,
such that $\mu_N < \theta < \mu_P$, then, as $n$ goes to infinity, the probability that a negative sample will have an
average response greater than $\theta$ (a false positive) goes to $0$.
This follows from Chebyshev's inequality.
By a similar argument, the chance of a false negative also goes to $0$ as
$n$ goes to infinity.

\noindent \textbf{Proposition 3:}
The weighted average $\mathbf{w}^T\mathbf{f}_i$
with smallest variance over positives (and negatives)
has equal weights.

\noindent \textbf{Proof:} We consider the case when $\mathbf{f}_i$'s are 
from positive
samples, the same being true for the negatives.
Then $\text{Var}(\sum_jw_j\mathbf{f}_i(j)/\sum_jw_j) = \sum w_j^2/(\sum w_j)^2 \sigma^2_P$.
We minimize $\sum w_j^2/(\sum w_j)^2$ by setting its partial
derivatives to zero and get $w_q(\sum w_j)=\sum w_j^2, \forall q$. Then
$w_q=w_j, \forall q,j$.

\chapter{Experimental analysis}
\label{cpt:experiments}
In this chapter we will present the features that we used, the ways in which we obtained them, the settings of our experiments and their corresponding results. 
In the first section we talk about the main datasets used in training and testing and the features created on them.
Then we present the experiments that we did and we analyse the results obtained. We compare our supervised and unsupervised approaches with regularized methods like Lasso and Elastic Net and with other well known methods used in feature
selection and classification: AdaBoost, SVM, FoBa. Our experiments refer to aspects like testing accuracy, training time, sensitivity to parameters and many others.

\section{Youtube-Objects experiments}

\subsection{Features design}

\begin {table}[H]
\begin{center}
\begin{tabular}{lcc}
\toprule
& Training dataset & Testing dataset \\
\midrule
No. of frames & 436970 & 134119 \\
No. of shots & 4200 & 1284 \\
No. of classes & 10 & 10 \\
\bottomrule
\end{tabular}
\caption{Youtube-Objects dataset statistics.}
\label{youtubeTable}
\end{center}
\end{table}

\paragraph{\textbf{Dataset description:}}
To train and test our system we used Youtube-Objects video dataset~\cite{youtube2012objects} and features obtained on ImageNet and CIFAR10.
%Youtube-Objects is the video dataset on which we tested our method and that was used in combination with other two image datasets to train the system.
%The video dataset that we used to test our approach and also for training in combination with other datasets is Youtube-Objects dataset \cite{youtube2012objects}. 
Details about Youtube-Objects are found in Table~\ref{youtubeTable}. The 10 classes are aeroplane,  bird,  boat,  car,  cat,  cow,  dog,  horse,  motorbike,  train.
This information refers to the entire training dataset, but in our experimental design, we used only a part of it to train the two methods. 
Details about the actual training set can be found in the next sections.
Each video in the dataset consists of a number of shots. The labeling is done per video; this means that some frames that appear in a 
video labeled as ``dog'' might contain only people and not dogs. This fact makes our task more difficult because we consider that some frames show
a certain object even though they do not.

\begin{figure}%[!ht]
\begin{center}
\includegraphics[scale = 0.9]{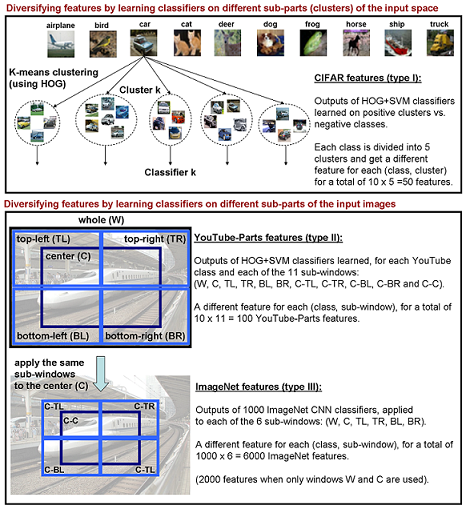}
\caption{Three types of features are created to form our pool. They are classifiers
trained on three different datasets: CIFAR10, Youtube-Objects and ImageNet. We encourage
feature diversity and independence by looking either at different parts of the input space
(type I) or at different locations in the image (types II and III). For type I we train separate classifiers
for each cluster of each class. For types II and III we train different classifiers for each class at every sub-window.
In total we get over 6160 features. As our experiments show the large variety of our features helps in
improving classification.
}
\label{fig:datasets}
\end{center}
\end{figure}

%In Figure~\ref{fig:datasets} we show some details about the datasets and the way in which the features were obtained.
\paragraph{\textbf{Features generation:}}
For the experiments we used a pool of 6160 features obtained on three different datasets, in different ways. The feature types
are the following:
\begin{enumerate}
 \item Features obtained by training binary classifiers on CIFAR10 dataset~\cite{CIFAR10}. These classifiers are trained on the data obtained by clustering the 
 images from each of the 10 classes into 5 clusters. The positive examples for each classifier are the examples from the corresponding class, 
 while negative examples are those from other classes (8 times more negatives than positives). These make a total of 50 features. The classes 
 from CIFAR10 coincide only partially (7 classes) with those from Youtube-Objects dataset. The classes from CIFAR10 are frog, truck, deer, 
 automobile, bird, horse, ship, cat,  dog, aeroplane.
 
 \item Features obtained by training a multiclass SVM classifier on the HOG applied on different parts of the frames. We have applied PCA on the 
 resulted HOG, thus we obtained smaller descriptors of length 46, so as to avoid as much as possible overfitting when using SVM. The training set that we 
 used to obtain these features is a subset of Youtube-Objects (25000 frames, equally distributed between the 10 classes). Then, we applied these 
 classifiers on each image and we considered as features the probabilities returned by SVM, this means 10 features for each part-classifier. 
 The parts of the images that we considered are the whole image, the center of the image (length and height are half the initial size), the four
 corners of the image, the center of the center of the image and the corners of the center of the image. Finally, we have 11 classifiers, each 
 with 10 probabilities, summing up to 110 features.
 
 \item Features obtained by using a pretrained network from Caffe~\cite{jia2014caffe}. The convolutional neural network was trained on the ImageNet dataset~\cite{imagenet} 
 and it contains 1000 features. We applied these features on our own dataset so: on the whole image, on the center and on the 4 
 corners of each example, thus we obtained $6000$ new features.
\end{enumerate}

\subsection{Experiments and results}
\label{sec:experiments}

We evaluate our method in the context of a limited training dataset and we intend to show that our method generalizes better than other well known methods. 
We combine features obtained from different image datasets and prove that knowledge transfer is useful by testing the system on
videos. In all the experiments we consider the accuracy per frame. In order to compare the different methods that we took into 
consideration, we varied some dimensions of the problem: the number of shots for training, the number of frames from each shot and the number 
of features considered. To avoid overfitting, besides studying the testing accuracy, we also observe training accuracy vs testing
accuracy.

\begin{figure}%[H]
\begin{center}
\includegraphics[scale = 0.8]{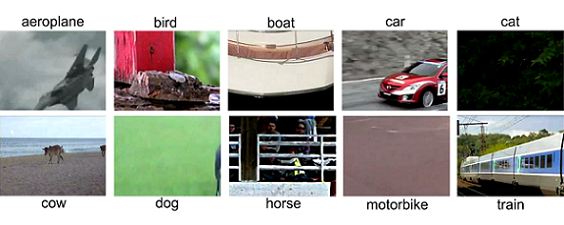}
\caption{One of the 30 random training sets used for one-shot learning - we use only one labeled frame per class. Note that some frames do not contain the object, making the learning task really difficult.}
\label{fig:one-shot-training}
\end{center}
\end{figure}

The Youtube-Objects dataset is a difficult one because the movies are taken in the wild and there are more categories of objects that appear simultaneously in a frame.
Moreover, in some frames of the videos the real object is even missing or other objects appear instead (e.g. a video is labeled as ''dog'', but it contains only a car in some of its frames). 
The shots in each video may differ very much. The differences are caused by the orientation, size,
luminosity, or by the presence of more object categories in the same frame. In some shots the object is occluded, it is in a corner 
or it is coming in and out. Due to these facts, learning a class of objects from a small number of frames becomes a very challenging
task. The splitting of the videos in training and testing is done as in~\cite{prest2012learning}. In Fig.~\ref{fig:one-shot-training} we show one of the random sets of training frames used in the case of \emph{one-shot learning}.
In this scenario we feed only one labeled frame from each class to the algorithm to learn the signs of the features.
You can notice that some of the frames are not at all representative for their category even though we chose the frame found in the middle of the random shot.
To ensure the accuracy of the results we have averaged the results of 30 or 100 random experiments for each method.

\begin{table}%[!ht]
\caption{The distribution of sub-windows for the input classifiers selected for
each category. The most selected one is \emph{whole}, indicating that
information from the whole image is relevant, including the background. For some classes
the object itself is also important as indicated by the high percentage of \emph{center} classifiers.
The presence of classifiers at other locations indicates that the object of interest, or objects
co-occurring with it might be present at different places. Note, for example the difference
in distribution between \emph{cats} and \emph{dogs}. The results suggest that looking more carefully
at location and moving towards detection, could benefit the overall classification.}
\label{distributionTable}
\begin{center}
\begin{tabular}{lcccccc}
\toprule   
Locations   & W       & C     & TL & TR & BL & BR  \\
\midrule
aeroplane   &  65.6    &   30.2    &   0     &    0          & 2.1           &   2.1   \\
bird        &  78.1    &   21.9    &   0     &    0          & 0               &   0       \\
boat        &  45.8    &   21.6    &   0     &    0          & 12.3          &   20.2  \\
car         &  54.1    &   40.2    &   2.0 &    0          & 3.7           &   0       \\
cat         &  76.4    &   17.3    &   5.0 &         0     & 1.3           &   0       \\
cow         &  70.8    &   22.2    &   1.8 &    2.4      &       0         &   2.8   \\
dog         &  92.8    &    6.2    &   1.0 &         0     &   0             & 0         \\
horse       &  75.9    &   14.7    &       0 &         0     &   8.3         & 1.2     \\
motorbike   &  65.3    &   33.7    &       0 &         0     &   0             & 1.0     \\
train       &  56.5    &   20.0    &       0 &    2.4      &   12.8        & 8.4     \\
\bottomrule
\end{tabular}
\end{center}
\end{table}

Regarding the type II of features we did an experiment to study whether there is a preference for features computed on a certain region. 
In Table~\ref{distributionTable} we show the distribution of the classifiers selected by our supervised method with respect to the position of the region on 
which they are obtained for each class. 
We can make some observations regarding this distribution. First, the whole image (W) is the most important for many classes, this means that apart
from the object, the environment is also important. Secondly, for some categories like car, motorbike, aeroplane the center (C) of the image is
important, while for others, regions off-center seem to be more representative than the center. Thirdly, for some classes that seem to be similar to 
humans, the classifiers chosen are rather different, as in the case of cats and dogs. For dogs the whole image is more representative, while for
cats different off-center regions are preferred. This might be due to the fact that cats can be found in more unconventional places than dogs that are bigger
and usually found on the ground.

We evaluated and compared eight methods: ours, SVM, AdaBoost, ours + SVM (feed to SVM only the features selected by our method), Lasso, Elastic Net, forward-backward 
selection (FoBa) and averaging. For SVM we used the most recent version (at the moment of the experiments) of LIBSVM~\cite{CC01a}, while for Lasso and Elastic Net we used the implementation provided in MATLAB. All the features have values between 0 and 1 and are expected to be positively correlated with the positive class.
In the case of our method, we select the features whose weights have a value greater than a threshold. After the features are selected, we can use
them with any classifier. The features are used with most of the tested algorithms exactly as they are, while with AdaBoost they should be transformed into classifiers, by
finding for each feature the threshold that minimizes the expected exponential loss at each iteration. This is the reason why AdaBoost proved to be much slower than the other methods.

We performed extensive experiments on both variants of our method (supervised and (almost) unsupervised) and also on the methods mentioned above. We evaluated: testing accuracy, 
training accuracy (to make sure the algorithm is not overfitting), training time, sensitivity to input parameters, accuracy of sign estimation, sparsity of the solutions and influence of the quantity of unlabeled data
over the recognition accuracy. In the majority of our experiments we are going to use four subsets of features: 1) all features of type I ($50$ features), 2) all features of types I and II ($160$ features), 3) $2000$ out of the $6000$ features of type III
- those computed on the whole image and on the central part of the image, 3) all features of types I and II and the $2000$ features of type III also selected in the previous case.

\paragraph{\textbf{Signs estimation:}}
The (almost) unsupervised setting supposes very limited labeled data, only for computing the signs of the features. It leads to very good performance even when it uses only one labeled frame per class to
flip the features. In Table~\ref{tab:sign_estimation} we show that the accuracy of signs estimation is usually high, it increases with the number of labeled shots and frames used. We can also notice an 
improvement in the sign estimation accuracy when the training sets contain stronger features like in the third and fourth case. The fact that the performance of our algorithm is good even for fewer labeled samples, as we will see in the next experiments, 
supports our affirmation that the algorithm is robust, not being sensitive to the signs of the features.

\begin{table}%[H]
\caption{Accuracy of feature sign estimation for different number of labeled shots: 
note that the signs are estimated mostly correctly even in the 1 labeled training 
frame per class case.
While not perfectly correct, the estimation agreements between signs estimated from a
single frame per class and signs estimated from the entire test set (containing more than $100K$ frames)
show that  estimating these feature signs from very limited labeled 
training data is feasible. Also, the experiments presented here indicate that our
approach is robust to sign estimation errors.}
\label{tab:sign_estimation}
\begin{center}
\begin{tabular}{lcccc}
\toprule
Number \\
of shots &   		Features~I           &  Features~I+II            &   Features~III         & Features~I+II+III          \\
\midrule
1   (1 frame)     	&  61.06\%          &  64.19\%                &  66.03\%          &  65.89\% \\
1   (10 frames)         &  62.61\%          &  65.64\%                &  67.53\%          &  67.39\%     \\
3   (10 frames)         &  66.53\%          &  69.31\%                &  73.21\%          &  72.92\%     \\
8   (10 frames)         &  72.17\%          &  73.36\%                &  78.33\%          &  77.97\%     \\
\hline
16  (10 frames)         &  74.83\%          &  75.44\%                &  79.97\%          &  79.63\%     \\
\hline
20  (10 frames)         &  75.51\%          &  76.23\%                &  80.54\%          &  80.22\%     \\
30  (10 frames)         &  76.60\%          &  77.15\%                &  80.98\%          &  80.70\%     \\
50  (10 frames)         &  77.41\%          &  77.79\%                &  81.52\%          &  81.24\%     \\

\bottomrule

\end{tabular}
\end{center}
\end{table}

\begin{figure}%[!ht]
\begin{center}
\includegraphics[scale = 0.44]{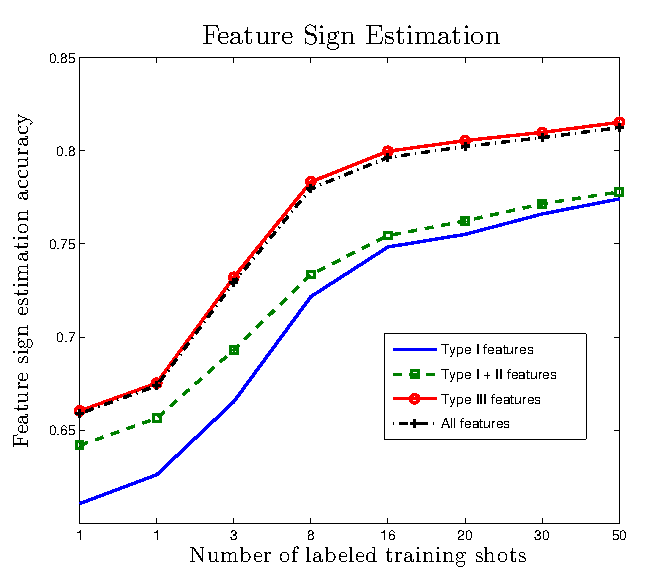}
\caption{
Accuracy of features signs estimation when the signs computed on the testing set are considered to be the ground-truth (the testing set size is bigger than $100K$ and it is more probable to predict the signs correctly when the samples are more numerous).
Note that the signs are better estimated when the features are stronger (as it happens in the case of type III and types I + II + III features), but the accuracy is quite high in all the four cases.
The first value corresponds to 1-shot-1-frame case (please also see Table~\ref{tab:sign_estimation}).
}
\label{fig:sign_estimation}
\end{center}
\end{figure}

\begin{table}[!ht]
\caption{Sign estimation accuracy for each class for the features selected by our algorithm when using 16 labeled shots with 10 labeled frames each. Results are averages of $30$ different experiments.}
\label{tab:signAccuracyPerClass}
\begin{center}
\begin{tabular}{lcccc}
\toprule
Locations   &   Feats.~I       &  Feats.~I+II         & Feats.~III       & Feats.~I+II+III          \\
\midrule
Aeroplane   &  99.00\%            &  100.00\%                &  100.00\%        &  100.00\% \\
Bird   &  93.17\%           &  85.33\%                &  64.83\%         &  63.87\%     \\
Boat  &  92.67\%           &  67.50\%                &  77.67\%         &  75.47\%     \\
Car  &  90.67\%          &  80.17\%                &  60.67\%          &  61.73\%     \\
Cat &  92.00\%          &  93.83\%                &  88.50\%          &  86.00\%     \\
Cow   &  90.50\%            &  92.83\%                &  70.17\%        &  74.67\% \\
Dog   &  91.33\%           &  96.50\%                &  89.33\%         &  88.47\%     \\
Horse  &  92.67\%           &  96.17\%                &  88.67\%         &  85.53\%     \\
Motorbike  &  90.17\%          &  87.00\%                &  71.83\%          &  78.33\%     \\
Train &  96.00\%          &  93.67\%                &  81.50\%          &  80.53\%     \\
\hline
Mean &  92.82\%          &  89.30\%                &  79.32\%          &  79.46\%     \\
\bottomrule
\end{tabular}
\end{center}
\end{table}

In 
%Table~\ref{tab:sign_estimation} and 
Fig.~\ref{fig:sign_estimation} we show the accuracy of the sign prediction for each feature, which depends only on the ground truth and the value of the features. We considered the ground truth for feature signs as being the signs obtained by
using the whole testing set as labeled data. In Table~\ref{tab:signAccuracyPerClass} we show how sign estimation accuracy varies for each class. In this case we computed the accuracy of the sign estimation particularly for the features selected 
with our unsupervised algorithm, not for all features. Both experiments were done on the four subsets of features described before. In the second experiment we studied only the case of 16 shots, each with 10 labeled frames per class, while in the first one we varied the number of shots and 
frames in order to see how it influences the sign accuracy. Notice (by comparing row 5 from Table~\ref{tab:sign_estimation} and 
last row from Table~\ref{tab:signAccuracyPerClass} which present same settings for the experiments) that our method chooses more features correctly flipped which means that the algorithm tends to select reliable features (the percent of the features selected
by the algorithm that have correct signs is higher than the percent of the total number of features that have correct signs).

\begin{figure}[ht]
\begin{center}
\includegraphics[scale = 0.6, angle = 0, clip]{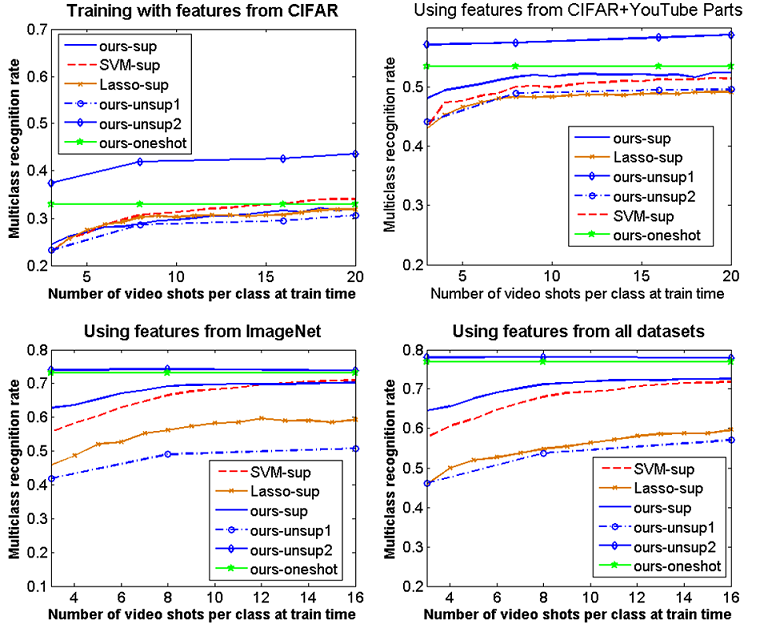}
\caption{Comparison of our unsupervised approach to the supervised case for different methods. In our case, unsup1 uses training data in
$J_u(\mathbf{w})$ only from the shots used 
by all supervised methods;
unsup2 also includes the testing data for learning, with unknown test labels; one-shot is unsup2 with a single labeled 
image per class used only for feature sign estimation. This truly demonstrates the ability of our approach to use minimum amount
of supervision and accurately learn in an unsupervised manner.}
\label{fig:semisup}
\end{center}
\end{figure}

\paragraph{\textbf{Methods comparison:}}
In Fig.~\ref{fig:semisup} we compare our supervised and unsupervised approaches with other methods used for feature selection. We notice that the results of ours-unsup2 are better than those for all the other methods.
Table~\ref{tab:lasso_elatic} shows the results obtained with regularized methods like: Lasso and Elastic Net compared to the results obtained with our supervised method. Notice how our algorithm performs much better when the number of
features is higher, while Elastic Net performs slightly better for the first set of features that are the least numerous. This suggests that our algorithm manages to better select from a higher number of features, which is quite encouraging 
because feature selection is more acutely needed when the number of features is very big. 
We tested Lasso and Elastic Net with different values of parameter $\lambda$ for each of the four subsets of features, and we chose the value for which we obtained the best results in the case of eight shots. 
The value of $\lambda$ is the same for Elastic Net and Lasso for the same subset of features.

\begin{table}[!ht]
\caption{Recognition accuracy for Lasso, Elastic Net and our supervised method.}
\centering
%\begin{minipage}{1.0\textwidth}\centering
\subfloat[Features of type I]{
\begin{tabular}{lccc}
\toprule
Feats.~I & Ours supervised & Lasso & Elastic Net ($\alpha=0.5$)\\%& Ridge regression \\
\midrule
1 shot &17.83\% & 17.70\% & \textbf{18.18\%}\\
3 shots	&21.98\% & 22.73\% & \textbf{22.76\%}\\
8 shots	&29.12\%& 30.22\% & \textbf{30.25\%}\\
16 shots & 29.75\%& 30.70\% & \textbf{30.72\%}\\
\bottomrule
\end{tabular}}
%\caption{Features of type I}
%\end{minipage}

%\centering
%\begin{minipage}{1.0\textwidth}\centering
\subfloat[Features of types I + II]{
%\subcaption{Type I + II}
\begin{tabular}{lccc}
\toprule
Feats.~I & Ours supervised & Lasso & Elastic Net ($\alpha=0.5$)\\%& Ridge regression \\
\midrule
1 shot &\textbf{36.71\%} & 32.67\% & 33.59\%\\
3 shots	&\textbf{46.63\%} & 42.89\% & 44.00\%\\
8 shots	&\textbf{51.06\%}& 48.33\% & 48.75\%\\
16 shots & \textbf{51.94\%}& 48.90\% & 49.02\%\\
\bottomrule
\end{tabular}}
%\caption{Features of types I + II}
%\end{minipage}

%\centering
%\begin{minipage}{1.0\textwidth}\centering
\subfloat[Features of type III]{
\begin{tabular}{lccc}
\toprule
Feats.~I & Ours supervised & Lasso & Elastic Net ($\alpha=0.5$)\\%& Ridge regression \\
\midrule
1 shot &\textbf{49.58\%} & 27.95\% & 28.31\%\\
3 shots	&\textbf{62.78\%} & 45.83\% & 46.37\%\\
8 shots	&\textbf{69.16\%}& 56.14\% & 56.39\%\\
16 shots & \textbf{70.23\%}& 59.31\% & 59.33\%\\
\bottomrule
\end{tabular}}
%\caption{Features of type III}
%\end{minipage}

%\centering
%\begin{minipage}{1.0\textwidth}\centering
\subfloat[Features of type I + II + III]{
\begin{tabular}{lccc}
\toprule
Feats.~I & Ours supervised & Lasso & Elastic Net ($\alpha=0.5$)\\%& Ridge regression \\
\midrule
1 shot & \textbf{52.39\%} & 27.57\% & 28.90\%\\
3 shots	&\textbf{64.52\%} & 45.81\% & 46.66\%\\
8 shots	&\textbf{71.21\%}& 54.84\% & 55.77\%\\
16 shots & \textbf{72.66\%}& 59.69\% & 60.72\%\\
\bottomrule
\end{tabular}}
%\caption{Features of type I + II + III}
%\end{minipage}
\label{tab:lasso_elatic}
%\label{fig:borrowed_signs}
\end{table}

\begin{figure}%[!t]
\begin{center}
\includegraphics[scale = 1.0]{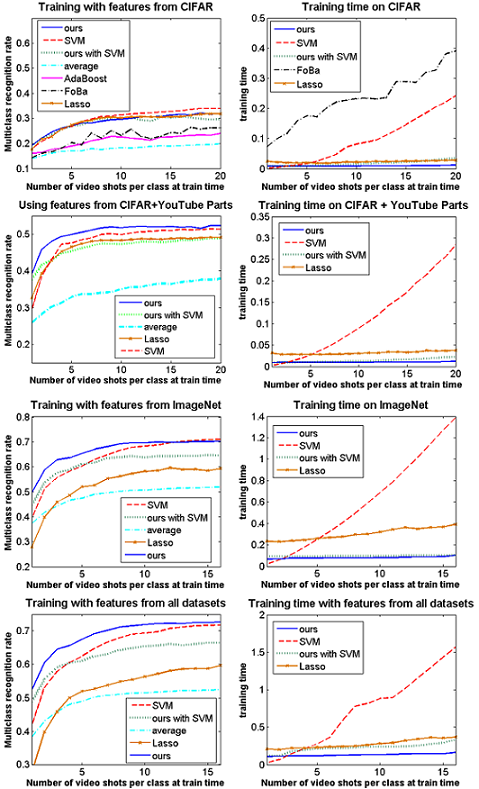}
\caption{Accuracy and training time on YouTube-Objects, with varying training video shots ($10$ frames per shot and results averaged over $30$ runs).
Input feature pool,
row~1: $50$ type~I features on CIFAR10;
row~2: $110$ type~II features on YouTube-Parts + $50$ type~I features on CIFAR10;
row~3: $2000$ type~III features in ImageNet;
row~4: $2160$ all features.
Ours outperforms SVM, Lasso, AdaBoost and FoBa.}

\label{fig:test_and_time}
\end{center}
\end{figure}

The results shown in Fig.~\ref{fig:test_and_time} prove that our supervised method is much faster than the other methods, it has a constant training time and 
also it outperforms them in accuracy, even SVM in many cases. The difference is more prominent when the number of shots is smaller. In the unsupervised case, we added \emph{unlabeled} training data.  Here, we outperformed 
all other methods by a very large margin, up to over $20\%$ (Table~\ref{tab:semisup} and Fig.~\ref{fig:semisup}). We tested with different amounts
of unlabeled data. While being almost insensitive to the number of labeled shots, (used only to estimate the feature signs), performance improved as more unlabeled data was added. Of particular note
is when only a single labeled image per class was used to estimate the feature signs, with all the other data being unlabeled (Fig.~\ref{fig:semisup}).
\begin{table}[!ht]
\caption{Experiments on our unsupervised learning. Improvement in recognition accuracy over our supervised method,
by using unsupervised learning with unlabeled test data.
Feature signs are learned using the same $(1,3,8,16)$ 
shots as in the supervised case. Note
that the first column presents the one-shot learning case, when
the unsupervised method uses a single labeled image per shot and
also per class. Results are averages over $30$ random runs.
}
\label{tab:semisup}
\begin{center}
\begin{tabular}{lcccc}  
\toprule
Training \# shots   &   1         &  3            & 8             & 16          \\
\midrule
Feature~I           &  +15.1\%   &  +15.2\%        &  +12.6\%     &  +12.6\%    \\
Feature~I+II        &  +16.7\%   &  +10.4\%        &  +6.4\%      &  +6.2\%     \\
Feature~III         &  +23.6\%   &  +11.3\%        &  +5.1\%      &  +3.7\%     \\
Feature~I+II+III    &  +24.4\%   &  +13.5\%        &  +6.9\%      &  +5.3\%     \\
\bottomrule
\end{tabular}
\end{center}
\end{table}

\begin{figure}[!ht]
\begin{center}
\includegraphics[scale = 0.7]{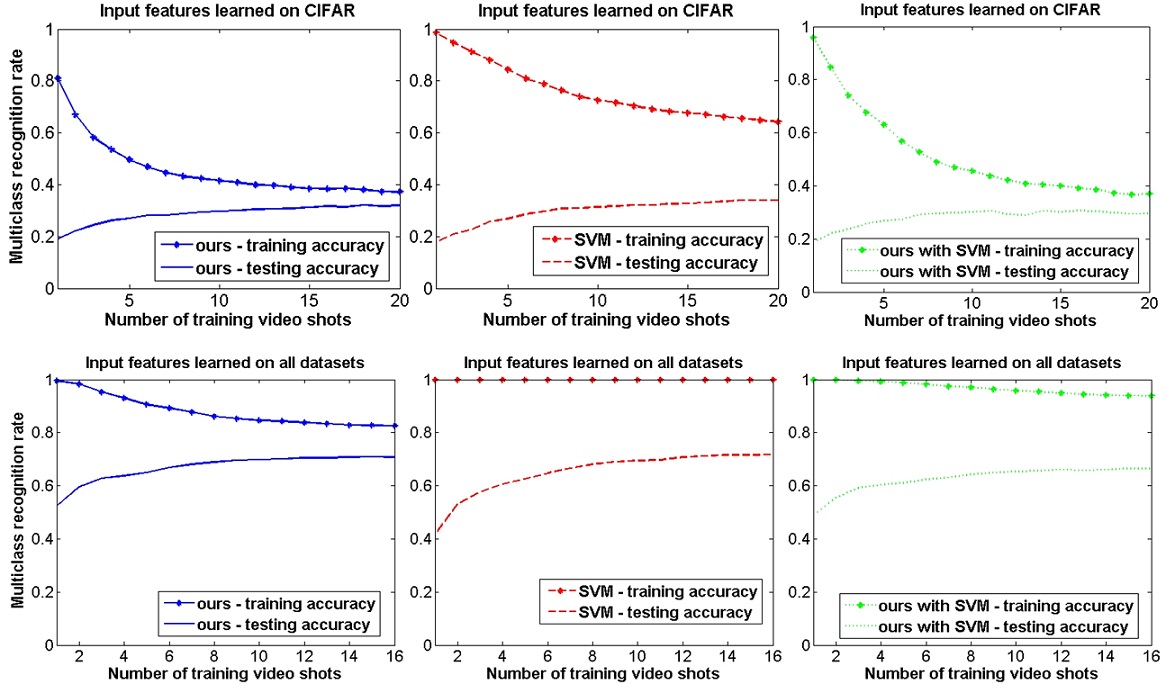}
\caption{Our method shows a good generalization power. It quickly learns subsets of features
that are more powerful on the testing set than when combined with SVM or SVM-alone. Note
that, in our case, the training and testing errors are relatively closer to each other
than for the competitors.
}
\label{fig:train_vs_test}
\end{center}
\end{figure}

\begin{table}%[H]
\caption{Recognition accuracy of our supervised method on Youtube-Objects dataset, using $10$ training shots (first row) and
$20$ training shots (second row), as we combine features from several datasets.
The accuracy increases significantly (it doubles) as the pool of features grows and becomes more diverse.
Features are:  type I: CIFAR; types I+II: CIFAR + Youtube-Parts; types I+II+III: CIFAR + Youtube-Parts + Imagenet (6000).
}
\label{tab:testing_nFeatures}
\begin{center}
\begin{tabular}{lccc}
\toprule   
Accuracy                 & Feats.~I ($50$)       & Feats.~I+II ($160$)    & Feats.~I+II+III ($6160$) \\
\midrule
$10$ train shots        &  29.69\%           &  51.57\%               &  69.99\%                      \\
$20$ train shots        &  31.97\%           &  52.37\%               &  71.31\%                      \\
\bottomrule
\end{tabular}
\end{center}
\end{table}

Our method also exhibits good generalization as we can notice in Fig.~\ref{fig:train_vs_test}, where training vs testing accuracy are plotted. This experiment is performed
on the supervised variant of our algorithm. We analysed the evolution of the testing accuracy with respect to the training accuracy in order to avoid overfitting. The size of 
the pool of features from which we choose a subset is very important when it comes to accuracy as we can see in Table~\ref{tab:testing_nFeatures}, this experiment
is also done on the supervised method. In this experiment we wanted to emphasize that the performance of our algorithm increases with the number of features used.

\begin{table}[!ht]
\caption{Mean accuracy per class, over 30 random runs of unsupervised learning with 16 labeled training shots.}
\label{tab:per_class_accuracy}
\begin{center}
\begin{tabular}{cccccccccc}
\toprule
\multicolumn{10}{c}{Mean accuracy per class (\%)}\\
\pbox{5cm}{Aero-\\plane} & Bird & Boat & Car & Cat & Cow & Dog & Horse & \pbox{5cm}{Motor-\\bike} & Train \\
\midrule
91.53 & 91.61& 99.11& 86.67& 70.02& 78.13& 61.63& 53.65& 72.66& 83.37 \\
\bottomrule
\end{tabular}
\end{center}
\end{table}

\begin{table}[!ht]
\caption{Recognition accuracy for our unsupervised method for four cases: 1) same frames used for unsupervised learning and for testing, 2) different frames for unsupervised learning and for testing, 3) different shots used for unsupervised learning and for testing and 
4) different videos used for unsupervised learning and for testing. In the first three cases we used the testing dataset for the unsupervised learning, while in the fourth case we used only the training set for the unsupervised learning, therefore the videos are 
different in training and testing.}
\label{tab:diff_shots_frames}
\centering
%\vspace{0.3cm}
%\begin{minipage}{1.0\textwidth}\centering
\subfloat[Features of type I]{
\begin{tabular}{lcccc}
\toprule
No. of shots & Same frames & Diff. frames & Diff. shots & Diff. videos\\%& Ridge regression \\
\midrule
%1 shot	&??\% & ??\% & ??\% & 26.89\%\\
3 shots	&37.24\% & 38.89\% & 27.18\% & 30.66\%\\
8 shots	&41.79\%& 43.11\% & 29.29\% & 34.21\%\\
16 shots &42.41\%& 43.79\% & 27.65\% & 33.54\%\\
\bottomrule
\end{tabular}}
%\subcaption{Features of type I}
%\end{minipage}
\vspace{0.3cm}
%\centering
%\begin{minipage}{1.0\textwidth}\centering
\subfloat[Features of types I + II]{
%\subcaption{Type I + II}
\begin{tabular}{lcccc}
\toprule
No. of shots & Same frames & Diff. frames & Diff. shots & Diff. videos\\%& Ridge regression \\
\midrule
%1 shot	&??\% & ??\% & ??\% & 48.97\% & 54.14\%\\
3 shots	&57.01\% &56.76\% & 52.96\% & 50.02\%\\
8 shots	&57.49\%& 57.29\% & 53.89\% & 50.07\%\\
16 shots &58.11\%& 58.03\% & 54.01\% & 50.05\%\\
\bottomrule
\end{tabular}}
%\subcaption{Features of types I + II}
%\end{minipage}
\vspace{0.3cm}
%\centering
%\begin{minipage}{1.0\textwidth}\centering
\subfloat[Features of type III]{
\begin{tabular}{lcccc}
\toprule
No. of shots & Same frames & Diff. frames & Diff. shots & Diff. videos\\%& Ridge regression \\
\midrule
%1 shot	&??\% & ??\% & ??\% & 26.94\% & 55.40\%\\
3 shots	&74.04\% & 73.95\% & 56.66\% & 55.60\%\\
8 shots	&74.25\%& 74.05\% & 57.86\% & 55.33\%\\
16 shots &73.87\%& 73.50\% & 58.29\% & 55.42\%\\
\bottomrule
\end{tabular}}
%\subcaption{Features of type III}
%\end{minipage}
\vspace{0.3cm}
%\centering
%\begin{minipage}{1.0\textwidth}\centering
\subfloat[Features of types I + II + III]{
\begin{tabular}{lcccc}
\toprule
No. of shots & Same frames & Diff. frames & Diff. shots & Diff. videos\\%& Ridge regression \\
\midrule
%1 shot	&??\% & ??\% & ??\% & 26.94\% & 62.03\%\\
3 shots	&78.02\% & 77.95\% & 61.24\% & 62.44\%\\
8 shots	&78.13\%& 77.96\% & 61.28\% & 62.11\%\\
16 shots &77.93\%& 77.79\% & 61.24\% & 62.11\%\\
\bottomrule
\end{tabular}
%\subcaption{Features of types I + II + III}
%\end{minipage}
}
\end{table}

\begin{figure}[!ht]
\begin{center}
\includegraphics[scale = 0.6, angle = 0, viewport = 0 0 400 300, clip]{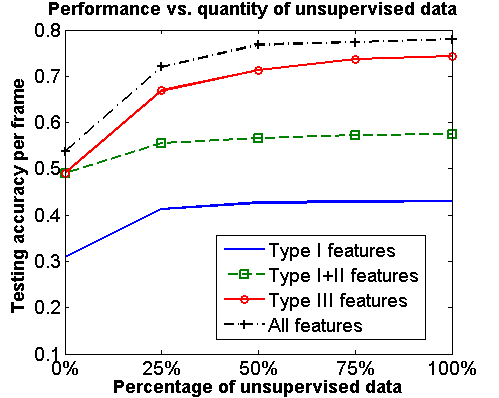}
\caption{Testing accuracy vs amount of unlabeled data used for unsupervised learning.
Note that the performance almost reaches a plateau after a relatively small fraction of unlabeled 
frames are added.
}
\label{fig:unsup_amount}
\end{center}
\end{figure}

\paragraph{\textbf{Unsupervised learning:}}
We also evaluate how testing accuracy is influenced by the quantity of unlabeled data used for learning. In order to assess this aspect, we trained our unsupervised
algorithm on different quantities of unlabeled data and we realized that, as expected, the accuracy increases with this quantity, but the variation is not so high. 
Once more than 25\% of the unsupervised data is provided, 
the accuracy reaches a plateau. These results are summarized in Table~\ref{tab:unsup_amount} and Fig.~\ref{fig:unsup_amount}.

\begin{table}%[!ht]
\caption{Testing accuracy when varying the quantity of unlabeled data used for unsupervised learning. We used $8$ video 
shots per class with 10 frames each, for estimating the signs of the features. Results are averages over $30$ random runs.}
\label{tab:unsup_amount}
\begin{center}
\begin{tabular}{lcccc}
\toprule
Unsupervised data   &   Feats.~I       &  Feats.~I+II         & Feats.~III       & Feats.~I+II+III          \\
\midrule
Train + 0\% test $S_A$   &  30.86\%            &  48.96\%                &  49.03\%        &  53.71\% \\
Train + 25\% test $S_A$   &  41.26\%           &  55.50\%                &  66.90\%         &  72.01\%     \\
Train + 50\% test $S_A$  &  42.72\%           &  56.66\%                &  71.31\%         &  76.78\%     \\
Train + 75\% test $S_A$  &  42.88\%          &  57.24\%                &  73.65\%          &  77.39\%     \\
Train + 100\% test $S_A$ &  43.00\%          &  57.44\%                &  74.30\%          &  78.05\%     \\
\bottomrule
\end{tabular}
\end{center}
\end{table}

In Table~\ref{tab:per_class_accuracy} we show how testing accuracy varies among classes. For this experiment we used 16 shots with 10 labeled frames per class to compute the signs of the 
features; we used features of types I, II and III. Note that the accuracy is generally higher for classes designing human-made objects like: boat, aeroplane, car, while for natural classes as dog, cat, horse the recognition accuracy is smaller. The difference
in accuracy might be explained by the fact that the videos for classes like boat, aeroplane and bird for which the accuracy is very high are rather static, the changes of the background and of the object itself are reduced,
while dogs, cats, and horses are more dynamic and the recognition task becomes more difficult. Especially for videos in classes aeroplane and boat the foreground is very uniform.

For the unsupervised case we also performed experiments in which the unlabeled training set contained distinct examples than those used for testing, because we wanted to see what is the influence of the unlabeled set on
the results. We considered four  cases that we present in Table~\ref{tab:diff_shots_frames}: 
1) same frames for unsupervised training and testing, 2) different frames for unsupervised training and testing, 3) different shots for unsupervised training and testing, 4) different videos for unsupervised training and testing. 
The results are as expected because the accuracy is higher when the frames used for the unsupervised learning are also used for testing, and it decreases
when the samples used for learning and testing are more distinct. The frames in the same shot are very similar to each other, and the shots within
the same video are more similar than the shots coming from different videos. However, the performance is still good even in the most difficult case, when
the examples used for training are actually extremely different from those used for testing.
%The accuracies in the first 2 cases are so close because the frames in a shot are very similar to each other.

\begin{figure}[!ht]

\begin{minipage}[b]{0.95\linewidth}
\centering
\includegraphics[width=\textwidth]{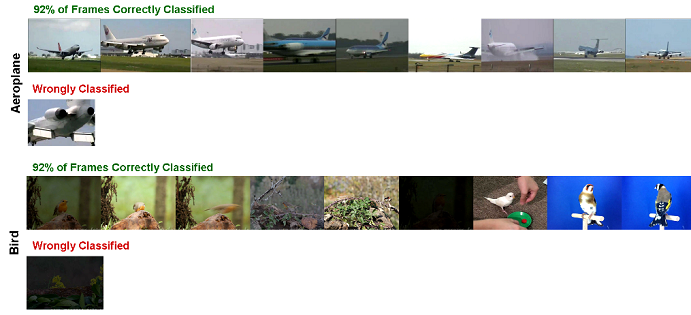}
\end{minipage}

\begin{minipage}[b]{0.95\linewidth}
\centering
\includegraphics[width=\textwidth]{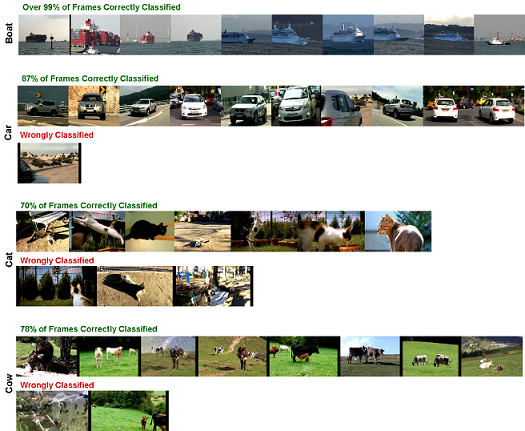}
\end{minipage}

\caption{
Lists of ten classified frames per category,
for which the ratio of correct to incorrect samples matches the mean class recognition accuracy.
}
\label{fig:examples_1_2}
\end{figure}

\begin{figure}[!ht]
\begin{center}
\includegraphics[width=\textwidth]{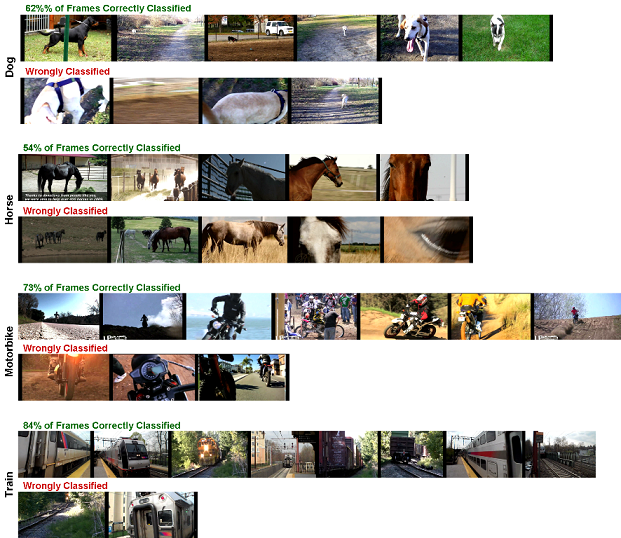}
\caption{Ten classified frames per class,
for which the ratio of correct to incorrect samples matches the mean class recognition accuracy.
}
\label{fig:examples_3}
\end{center}
\end{figure}

For a better understanding of the performance of our unsupervised method, we present in Figs.~\ref{fig:examples_1_2} and~\ref{fig:examples_3} for each of the 10 categories of objects images that were classified correctly and incorrectly. For each class the proportion of
correct and incorrect examples is consistent with the recognition accuracy per class. These results are obtained when testing our one-shot learning algorithm (only one labeled example is used per class). Note that we considered all frames in a 
video shot as belonging to a single category - even though sometimes a significant amount of frames did not contain any of the above categories. Therefore, often our results look qualitatively better than the quantitative evaluation.

\paragraph{\textbf{Signs transfer:}}
Another idea that we investigated during our experiments was the possibility to transfer the signs from a category to another. We computed the signs of the features for class \emph{cat} and used them also for class
\emph{dog}. This would be a very useful idea if we have some classifiers already learnt and we want to learn a new category for which we do not have labeled images. Then we can take the signs from one of the 
classifiers and use them for the new class. We made two experiments. In the first one we
computed the binary accuracy for each class individually for three distinct cases: 1) the signs used were the real signs, 2) the signs were taken from a very similar category, 3) the signs were taken from a very dissimilar category (the evaluation of
the similarity/dissimilarity of the classes is decided by us, so it might be subjective). We can notice a decrease in accuracy when the signs used were not the original ones, and the decrease is more pronounced when borrowing the signs from more dissimilar classes.
In Table~\ref{tab:bin_acc_sign_transf} we present the results for two sets of features: for types I + II and for types I + II + III, while in Table~\ref{tab:sign_transf_classes_a} 
we show the classes from which we borrowed the signs.

\begin{table}
\caption{Mean binary accuracy per class, over 30 random runs of unsupervised learning with 8 labeled training shots. We tested three cases: 1) the signs are computed on each class, 2) the signs are borrowed from
another very similar class, 3) the signs are borrowed from a very dissimilar class. Results are obtained on two subsets of features. See also Table~\ref{tab:sign_transf_classes}a for the classes considered similar~/~dissimilar.}
\label{tab:bin_acc_sign_transf}
%\begin{center}
\centering
\subfloat[Features of types I + II]{ 

\begin{tabular}{l|ccc}
\toprule
&\multicolumn{3}{c}{Testing accuracy(\%)}\\
\midrule
\pbox{5cm}{Class\\name} & \pbox{5cm}{Its\\own\\sign} &\pbox{5cm}{From\\sim.\\class} & \pbox{5cm}{From\\dissim.\\class}\\
\midrule
Aeroplane & 96.85 & 84.93 & 87.77\\
Bird	& 96.73 & 94.53 & 94.58\\
Boat	& 98.34 & 88.34 & 87.50\\
Car	& 97.53 & 96.90 & 96.81\\
Cat 	& 92.30 & 92.23 & 92.08\\
Cow	& 94.16 & 94.14 & 94.13\\
Dog	& 80.79 & 80.73 & 81.26\\
Horse	& 88.95 & 88.99 &80.72\\
Motorbike & 95.10 & 95.10 &95.10\\
Train 	& 93.35 & 87.33 &87.06\\
\midrule
Mean 	& 93.41 & 90.32&89.70\\
\bottomrule
\end{tabular}}
%\end{minipage}%
\hspace{0.5cm}%
\subfloat[Features of types I + II + III.]{
%\centering
%\subcaption{Features of types I + II + III.}
\begin{tabular}{l|ccc}
\toprule
&\multicolumn{3}{c}{Testing accuracy(\%)}\\
\midrule
\pbox{5cm}{Class\\name}&\pbox{5cm}{Its\\own\\sign} & \pbox{5cm}{From\\sim.\\class} & \pbox{5cm}{From\\dissim.\\class}\\
\midrule
Aeroplane & 96.63 & 91.73 & 93.01\\
Bird	& 99.01 & 97.68 & 94.59\\
Boat	& 97.43 & 98.29 & 93.25\\
Car	& 98.67 & 98.47 & 97.73\\
Cat 	& 94.48 & 94.66 & 93.12\\
Cow	& 95.80 & 95.87 & 94.21\\
Dog	& 84.87 & 82.13 & 80.88\\
Horse	& 92.05 & 91.03 &81.26\\
Motorbike & 98.09 & 97.67 &97.19\\
Train 	& 94.28 & 92.26 &87.82\\
\midrule
Mean 	& 95.13 & 93.97&91.30\\
\bottomrule
\end{tabular}
%\end{minipage}
}
\end{table}

\begin{table}
\caption{The classes from which we borrowed the signs in the experiments with the sign transfer.}
\label{tab:sign_transf_classes}
\centering
%\begin{minipage}{0.45\linewidth}
\label{tab:sign_transf_classes_a}
\subfloat[The classes that we considered similar~/~dissimilar in order to borrow the feature signs from.]{
%\centering
%\subcaption{The classes that we considered similar~/~dissimilar in order to borrow the feature signs from.}
\begin{tabular}{lcc}
\toprule
\pbox{5cm}{Class\\name} & \pbox{5cm}{Very sim.\\class} & \pbox{5cm}{Very dissim.\\class}\\
\midrule
Aeroplane & Bird & Train\\
Bird	& Aeroplane & Train \\
Boat	& Aeroplane & Cat \\
Car	& Motorbike & Bird \\
Cat 	& Dog & Boat \\
Cow	& Horse & Cat \\
Dog	& Cat & Boat \\
Horse	& Cow & Aeroplane \\
Motorbike & Car & Cat\\
Train 	& Car & Cow \\
\bottomrule
\end{tabular}}
%\end{center}
%\end{table}
%\end{minipage}%
\hspace{0.5cm}%
%\begin{minipage}{0.45\linewidth}
\subfloat[Classes chosen to borrow the feature signs from for the experiments in which we kept 6 original signs and 4 original signs.]{
\label{tab:sign_transf_classes_b}
%\centering
%\subcaption{Classes chosen to borrow the feature signs from for the experiments in which we kept 6 original signs and 4 original signs.}
%\label{tab:multiclass_transf_signs}
\begin{tabular}{lcc}
\toprule
\pbox{5cm}{Class\\name} & \pbox{5cm}{6 orig.\\signs} & \pbox{5cm}{4 orig.\\signs}\\
\midrule
Aeroplane & Aeroplane & Aeroplane\\
Bird	& Bird & Aeroplane \\
Boat	& Aeroplane & Aeroplane \\
Car	& Car & Car \\
Cat 	& Cat & Cat \\
Cow	& Cow & Horse \\
Dog	& Cat & Cat \\
Horse	& Cow & Horse \\
Motorbike & Car & Horse\\
Train 	& Train & Car \\
\bottomrule
\end{tabular}}
%\end{minipage}
%\label{fig:classes_signs}
\end{table}

In the second experiment we evaluated the multiclass accuracy in 3 different cases: 1) when all the signs were the original ones, 2) the signs were
the original ones for 6 classes, while for the other 4 they were borrowed, 3) the signs for 4 classes were the original ones, while for the other 6 classes they
were borrowed. The results are summarized in Table~\ref{tab:borrowed_signs}. We can notice that the accuracy generally decreases when the signs are borrowed and we
do not use all the original ones.
%Tables~\ref{tab:multiclass_acc_CIFAR},~\ref{tab:multiclass_acc_CIFAR_parts},~\ref{tab:multiclass_acc_2000} and~\ref{tab:multiclass_acc_2160}.
For the two new settings of the experiments, for each of the classes we present in Table~\ref{tab:sign_transf_classes_b} the classes
on which we computed their signs.

\begin{table}
\caption{
Multiclass accuracy for 3 settings: 1) with the original signs, 2) with 6 original signs, 3) with 4 original signs.
See also Table~\ref{tab:sign_transf_classes}b for the classes from which the signs are borrowed.
}
\centering
%\begin{minipage}{0.45\textwidth}\centering
\subfloat[Features of type I]{
\begin{tabular}{lccc}
\toprule
&\multicolumn{3}{c}{Multiclass accuracy}\\
\pbox{5cm}{No. of\\labeled\\shots} & \pbox{5cm}{All orig.\\signs} &\pbox{5cm}{6 orig.\\signs} & \pbox{5cm}{4 orig.\\signs}\\
\midrule
1 & 34.67\% & 35.14\% & 34.28\%\\
3& 38.79\% & 37.37\% & 34.06\%\\
8& 43.06\% & 40.05\% & 34.91\%\\
16 & 43.67\% & 40.45\% & 35.38\% \\
\bottomrule
\end{tabular}}
%\subcaption{Features of type I}
%\end{minipage}%
\hspace{0.5cm}%
%\begin{minipage}[h]{0.5\linewidth}\centering
\subfloat[Features of types I + II]{
\begin{tabular}{lccc}
\toprule
&\multicolumn{3}{c}{Multiclass accuracy}\\
\pbox{5cm}{No. of\\labeled\\shots} & \pbox{5cm}{All orig.\\signs} &\pbox{5cm}{6 orig.\\signs} & \pbox{5cm}{4 orig.\\signs}\\
\midrule
1 & 54.95\% & 54.14\% & 51.40\%\\
3& 57.01\% & 55.67\% & 51.02\%\\
8& 57.49\% & 55.25\% & 50.32\%\\
16 & 58.11\% & 55.77\% & 50.25\% \\
\bottomrule
\end{tabular}}
%\subcaption{Features of types I + II}
%\end{minipage}
\vspace{0.5cm}

\subfloat[Features of type III]{
%\begin{minipage}{0.45\textwidth}\centering
\begin{tabular}{lccc}
\toprule
&\multicolumn{3}{c}{Multiclass accuracy}\\
\pbox{5cm}{No. of\\labeled\\shots} & \pbox{5cm}{All orig.\\signs} &\pbox{5cm}{6 orig.\\signs} & \pbox{5cm}{4 orig.\\signs}\\
\midrule
1 & 73.52\% & 72.46\% & 72.66\%\\
3& 74.04\% & 72.46\% & 71.96\%\\
8& 74.25\% & 72.12\% & 73.32\%\\
16 & 73.87\% & 72.38\% & 73.08\% \\
\bottomrule
\end{tabular}}
%\subcaption{Features of type III}
%\end{minipage}%
\hspace{0.5cm}%
%\begin{minipage}[h]{0.5\linewidth}\centering
\subfloat[Features of types I + II + III]{
\begin{tabular}{lccc}
\toprule
&\multicolumn{3}{c}{Multiclass accuracy}\\
\pbox{5cm}{No. of\\labeled\\shots} & \pbox{5cm}{All orig.\\signs} &\pbox{5cm}{6 orig.\\signs} & \pbox{5cm}{4 orig.\\signs}\\
\midrule
1 & 76.86\% & 76.08\% & 75.95\%\\
3& 78.02\% & 76.01\% & 75.61\%\\
8& 78.13\% & 76.34\% & 76.19\%\\
16 & 77.93\% & 76.29\% & 75.93\% \\
\bottomrule
\end{tabular}}%
%\subcaption{Features of types I + II + III}
%\end{minipage}%

\label{tab:borrowed_signs}
\end{table}

\begin{figure}[!ht]
\begin{minipage}[b]{0.495\linewidth}
\centering
\includegraphics[width=\textwidth]{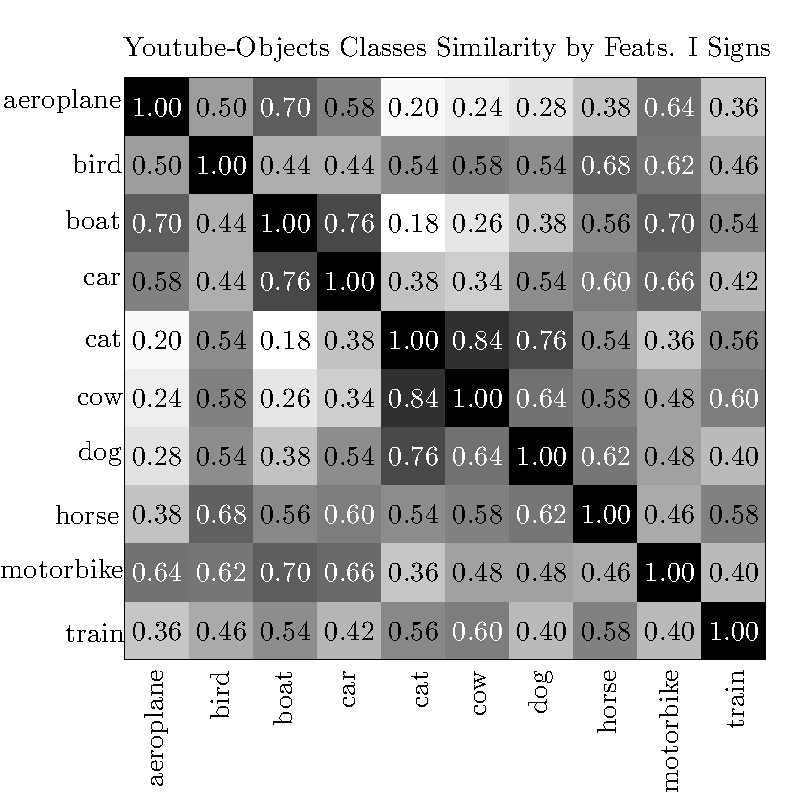}
\end{minipage}
\begin{minipage}[b]{0.495\linewidth}
\centering
\includegraphics[width=\textwidth]{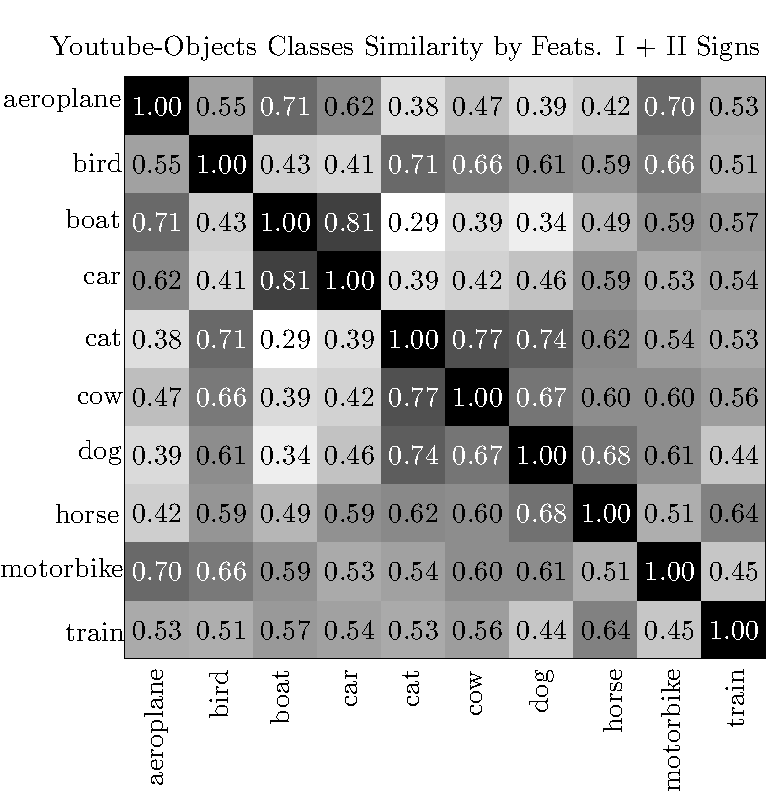}
\end{minipage}
\begin{minipage}[b]{0.49\linewidth}
\centering
\includegraphics[width=\textwidth]{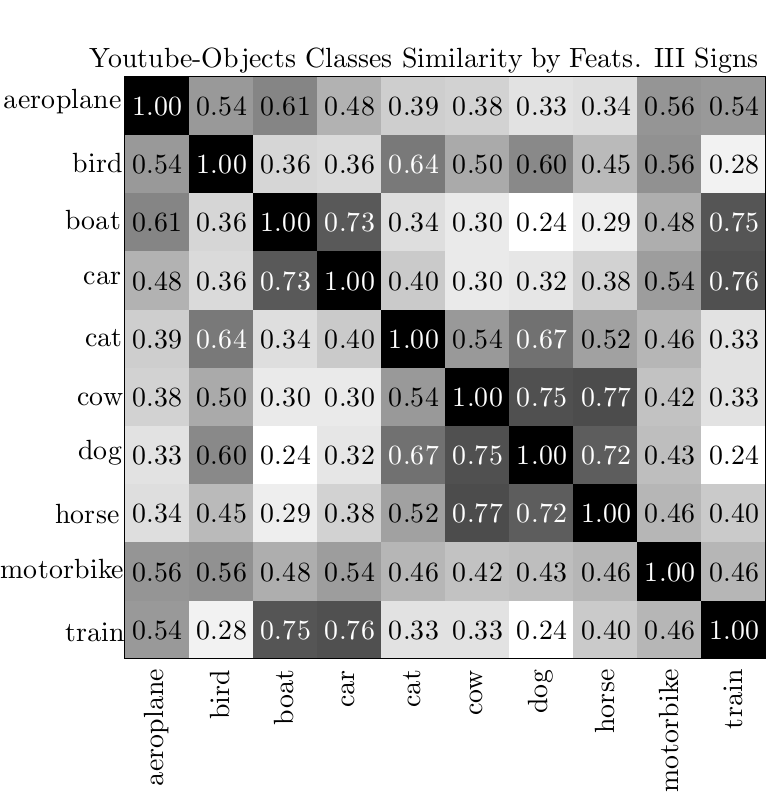}
\end{minipage}
\begin{minipage}[b]{0.5\linewidth}
\centering
\includegraphics[width=\textwidth]{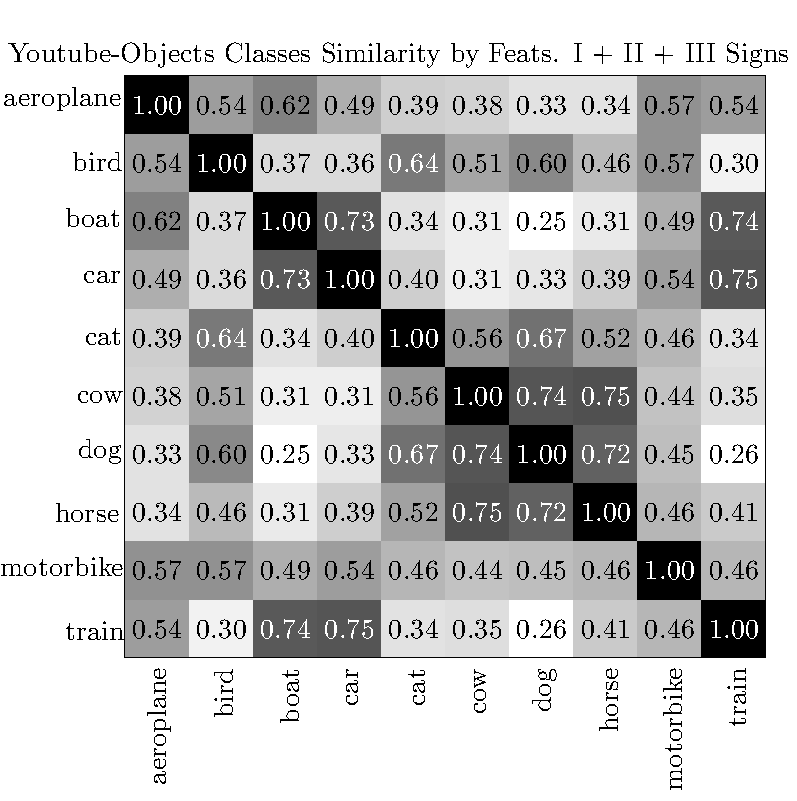}
\end{minipage}
\caption{Youtube-Objects classes similarity based on the signs of the features. For each pair of features we present the percent of signs of features that coincide. The higher this percent is, the higher the similarity is.}
\label{fig:class_similarity_by_signs}
\end{figure}

Another interesting experiment related to the sign transfer was to evaluate the similarity/dissimilarity of the classes based on the percent of feature 
signs that coincide for each pair of classes. We try to find a more objective criterion (a numerical one) in order to decide which classes are similar and which are not. In Fig.~\ref{fig:class_similarity_by_signs} we show for each class how similar it is to all classes
in the dataset. We can notice that the similarities computed in this way are quite intuitive and the more and stronger features we have, the more intuitive the similarities found are. 
The similarities computed with the fourth subset of features (types I + II + III) are better than those found only with features of type I. Let us focus on the last case considered and look at the classes to which
the similarity is~$> 0.5$; for \emph{aeroplane: boat, motorbike, bird, train}, for \emph{bird: cat, dog, motorbike, aeroplane, cow}, for \emph{boat: train, car aeroplane}, for \emph{car: train, boat, motorbike}, for \emph{cat: dog, bird, cow, horse}, 
for \emph{cow: horse, dog, cat, bird}, for \emph{dog: cow, horse, cat, bird}, for \emph{horse: cow, dog, cat}, for \emph{motorbike: aeroplane, bird, car}, for \emph{train: car, boat, aeroplane}. We can remark the fact that 
generally the classes that designate animals are similar to each other, while classes related to transportation (which are also human-made) are more similar between them according to the signs of the features. This result is not at 
all surprising, we expected that the categories that are semantically related to be more similar than those that are not. It would have been counterintuitive to obtain that the \emph{train} is similar to the \emph{cat}, for example.

\section{MNIST experiments}

\begin{figure}%[H]
\begin{center}
\includegraphics[scale = 0.40]{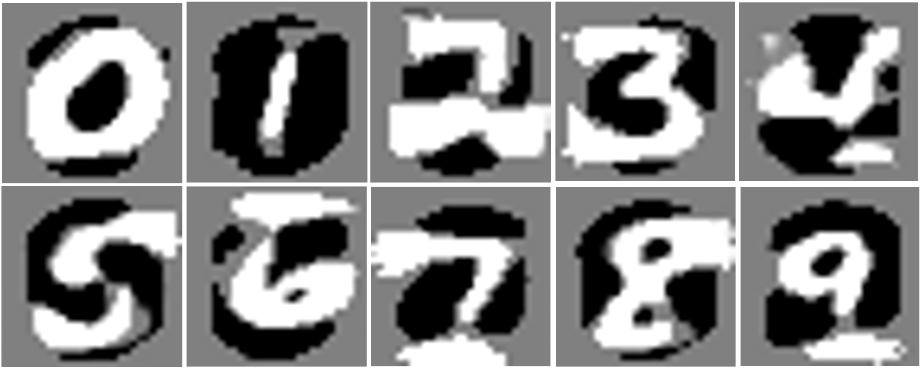}
\caption{We represented the classifiers chosen in white and black: white for the positively correlated, black for negatively correlated and grey for those not chosen.}
\label{fig:MNIST_classifs}
\end{center}
\end{figure}

In order to assess more accurately the performance of the algorithm that we developed we have tested it on MNIST dataset (containing images with digits). For these experiments we made a small change to the algorithm. 
The data are normalized so that they have the mean equal to 0 and the standard deviation equal to 1. Therefore, the values are not anymore between 0 and 1, they might also be negative and not subunitary. We noticed that when we
flip the features it would be better to use $-f$ instead of $1 - f$ as we did before. This new way of flipping the features is used in the MNIST experiments.
We show in Fig.~\ref{fig:MNIST_classifs} the classifiers chosen for each class. We represented
in black the negatively correlated features (before flipping, because after flipping all features are positively correlated), in white the positively correlated features (before flipping) and in grey the features that 
were not chosen. The number of classifiers chosen was k = 400. The number of labeled images per class used for learning the signs was 2000.
We can notice that the classifiers chosen are those from the center of the image. The positively correlated ones are precisely those 
that represent the shape of the digit, while the negatively correlated are around them and emphasize the shape of the digit. 

\begin{table}[!ht]
\caption{Mean multiclass accuracy for unsupervised learning on MNIST dataset for 30 random experiments. We varied the number of labeled images on which we learnt
the signs, and we used the whole testing set for unsupervised learning.}
\label{tab:multiclass_MNIST}
\begin{center}
\begin{tabular}{lcc}
\toprule
%\backslashbox{No. of labeled images}{Accuracy} & Whole image & Center\\% & Using training set only\\
No. of labeled images & Whole image & Center\\% & Using training set only\\
\midrule
1	& 64.36\% & 32.93\%\\% & 62.79\%\\
8	& 74.75\% & 58.88\%\\% & 74.01\%\\
16	& 75.70\% & 62.26\%\\%& 74.90\%\\
64	& 76.57\% & 65.89\%\\%& 75.96\%\\
128	& 76.63\% & 66.51\%\\%& 76.00\%\\
512	& 76.63\% & 67.19\%\\%& 75.98\%\\
1024	& 76.61\% & 66.90\%\\%& 75.89\%\\
2048	& 76.68\% & 67.07\%\\%& 75.86\%\\
6000 	& 76.57\% & 67.80\%\\%& 75.83\%\\
\bottomrule
\end{tabular}
\end{center}
\end{table}

\begin{figure}[!ht]
\begin{center}
\includegraphics[scale = 0.50]{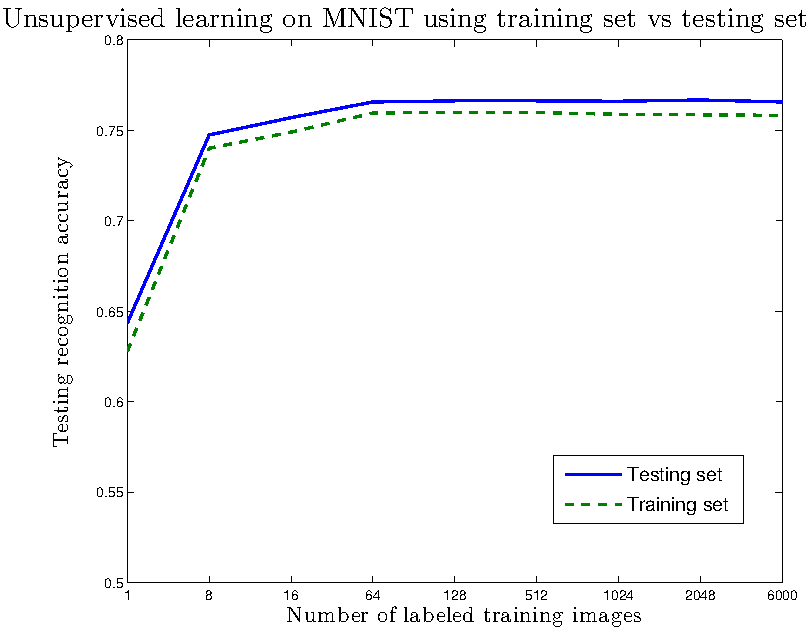}
\caption{Recognition accuracy for the unsupervised learning algorithm when: a) the testing set is used for unsupervised learning, b) the training set only is used for the
unsupervised learning.}
\label{fig:train_test_unsup_MINST}
\end{center}
\end{figure}

The multiclass
recognition accuracies obtained by our algorithm on the MNIST dataset are found in Table~\ref{tab:multiclass_MNIST}. We learnt the signs of the features on different numbers of images per class, ranging from
one image per class, up to all (around 6000) images per class. We also present the results obtained when we use instead of the whole image, only its central part. Even though the number of features is 
halved in this case, the accuracy when the center is used nears the accuracy obtained with the whole image when the number of labeled examples increases, although 
for 1 labeled image the accuracy when only the center is used is half the accuracy with the whole image.

\begin{figure}[!ht]
\begin{center}
%[scale = 0.45, angle = 0, viewport = 0 0 500 400, clip]
\includegraphics[scale = 0.45]{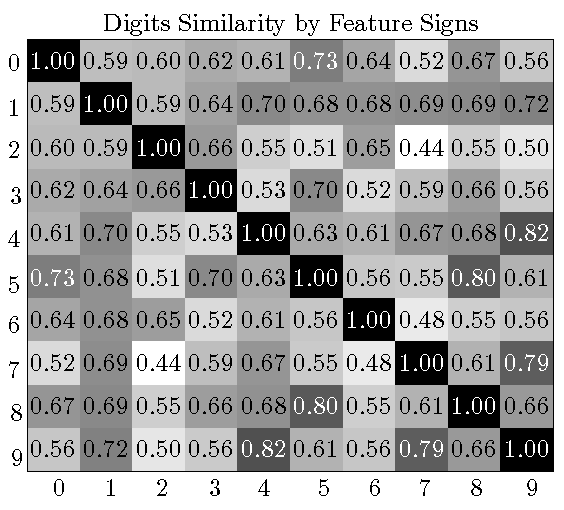}
\caption{Digits similarity in MNIST dataset based on the percent of feature signs that coincide among them.}
\label{fig:digits_sign_similarity}
\end{center}
\end{figure}

In Fig.~\ref{fig:train_test_unsup_MINST} we showed the testing accuracy of our unsupervised algorithm for two different settings: 1) the unsupervised learning was done on the testing set, 2) the unsupervised learning was
done on the training set. We can notice that the difference in accuracy between the two is extremely small, this means that the algorithm can generalize very well
and the power of the classification method does not necessarily come from the testing examples used during the unsupervised learning.
We have also performed an experiment that assesses the similarity between the ten classes (digits) by evaluating the percent of signs that
coincide between each pair of classes. In Fig.~\ref{fig:digits_sign_similarity} we show the level of the sign coincidence for each pair of digits.

\chapter{Conclusions and future work}
\label{cpt:conclusion}
These last sections are reserved to the final discussions regarding the results presented in the previous chapter and the whole method. We will make a higher level analysis of the results. 
Then we draw some conclusions about our entire work, and finally, we present briefly how we intend to continue our research.
\section{Discussion}

\paragraph{\textbf{Discussion on (almost) unsupervised learning:}} 

We demonstrated that our approach is able 
to learn superior classifiers in the case when no data labels are available,
but only the signs of features 
are known. In our experiments, we only used minimal data to estimate these signs. Once they are known, any amount of unlabeled data can be incorporated. This aspect reveals a key insight:
being able to label \emph{the features}, and \emph{not the data}, is sufficient for learning.  For example, when learning to separate oranges from cucumbers, 
if we knew the features that are positively correlated with the ``orange'' class (roundness, redness, sweetness, temperature, latitude where image was taken) in the immense
sea of potential cues, we could then employ huge amounts of unlabeled images
of oranges and cucumbers, to find the best relatively small group of such features. Also note that since only a small number of images are used
for estimating the feature signs (as few as one per class), some signs may be wrong.
However, the very weak sensitivity of the method to the number of labeled training samples strongly indicates that it is robust to noise in sign estimation, 
as long as most of the features are correctly oriented.

\paragraph{\textbf{Discussion on the selected features:}}

\begin{figure}
\centering
\includegraphics[scale = 1.0]{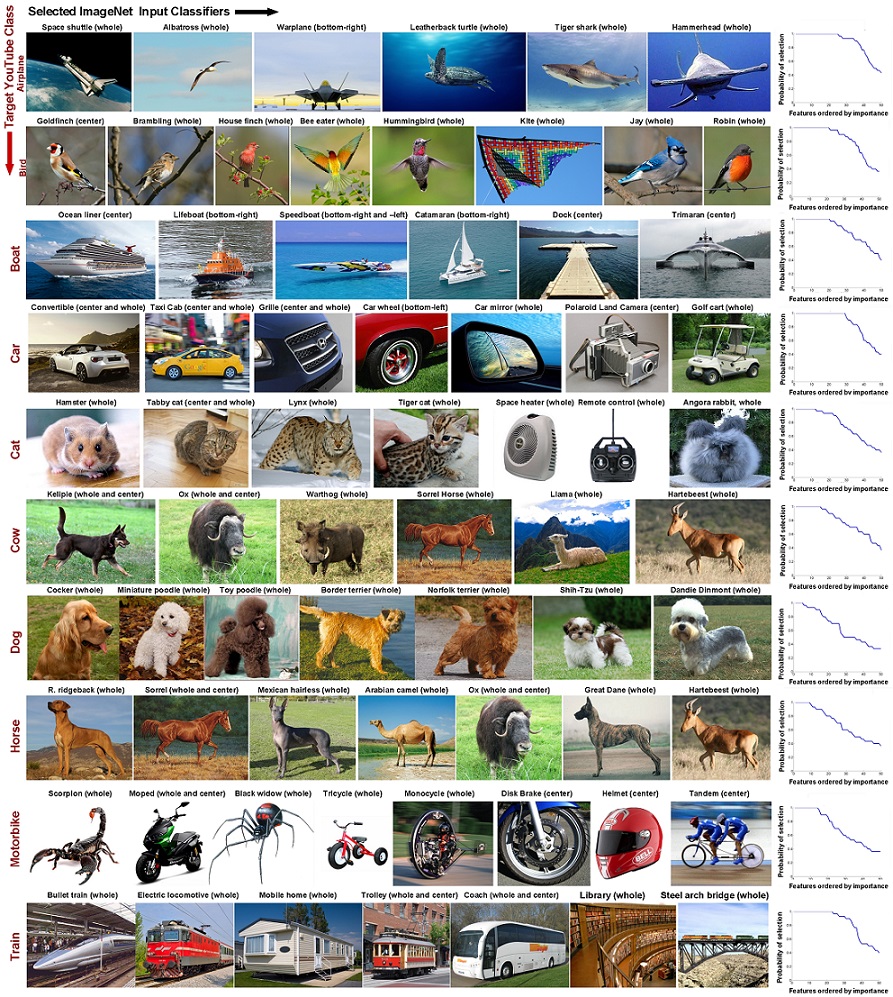}
\caption{For each training target class from Youtube-Objects videos (labels on the left), we present the most frequently selected
ImageNet classifiers (input features), over $30$ independent experiments, with $10$ frames per shot and
$10$ random shots for training. In images we show the classes that were always selected by our method when $k=50$.
On the right we show the probability
of selection for the most important $50$ features together with other relevant classes and their frequency of selection presented as a list.
Note how stable the selection process is and how related (but not identical) the selected
classes are in terms of appearance, context or geometric part-whole relationships,
to work robustly together as an ensemble. We find two aspects indeed surprising: 1) the high probability (perfect 1) of selection
of the same classes, even for such small random training sets and 2) the fact that unrelated classes, in terms of meaning,
could be so useful for classification, based on their surprising shape and appearance similarity.}
\label{fig:YoutubeImageNet}
\end{figure}

We have noticed some surprising ways in which the class of a frame in Youtube-Objects is associated with a series of classes in ImageNet. There
are different ways in which these associations are done:
\begin{enumerate}
  \item similarity of the global appearance of the two objects, but no semantic relation: eg. train vs banister, tigershark vs. plane, Polaroid 
camera vs. car, scorpion vs. motorbike, remote control vs. cat's face, space heater vs. cat's head.
  \item co-occurrence and similar context: helmet vs. motorbike
  \item part-to-whole object relation: grille, mirror and wheel vs. car
  \item combinations of the previous: dock vs. boat, steel bridge vs. train, albatross vs. plane.
\end{enumerate}

Another observation would be the fact that some of the classes play a role of borders between the positive class and the others. This ensures the
separation between the main class and the neighbouring classes. 
Another benefit is the fact that although there is no classifier for a certain class, it manages to learn how to distinguish this class from the
others by using together other existent classes that are similar to it. For example, even though in ImageNet there is not a ``cow'' class, it learns
the new concept from the ones that are available.
In order to support our claims we show in Figure~\ref{fig:YoutubeImageNet} for each class in Youtube-Objects the classes from ImageNet whose weights were the biggest, 
which means that they mattered more. We can notice that many selected classes are similar in appearance to the positive class, this is most visible in the case of the 
aeroplane class, while for other classes the resemblance is also at the semantic level, not only in appearance.

\section{Conclusion}

We present a fast feature selection and learning method that requires minimal supervision, with strong theoretical properties and excellent generalization and accuracy in practice.
The crux of our approach is its ability to learn from unlabeled data once the feature signs are determined. Our contribution could open doors for new and exciting research in machine 
learning, with practical and theoretical impact. Both our supervised and unsupervised approaches can 
quickly learn from limited data and identify sparse combinations of features that outperform powerful methods such as SVM, AdaBoost, Lasso and greedy sequential selection --- in both time and accuracy.
With a formulation that permits very fast optimization and effective learning from large heterogeneous feature pools, our approach provides a useful tool for many recognition tasks, 
suited for real-time, dynamic environments. Our work complements much of the machine learning research on developing new, more powerful, classifiers.
While this thesis has primarily demonstrated the effectiveness of our feature combinations in a specific context, our methods are general and could be used in conjunction with any machine learning algorithm.

We tested the method on a difficult video dataset and also showed that knowledge transfer is possible between datasets with very different characteristics: starting from different object classes, to different image quality and positioning of the target object.
The method needs very limited labeled data for computing the signs of the features (whether they are positively or negatively correlated). It manages to compute the signs quite well even when only one frame per class is presented. And the method can
handle successfully high quantities of unlabeled data. Moreover, after a percent of the unlabeled data are presented to the algorithm, the recognition accuracy reaches a plateau which means that even fewer examples are enough to learn.
Either the supervised and the unsupervised approaches are better than most of the methods mentioned above.

The proposed method has strong theoretical properties; it guarantees the sparsity of the solution, all features have the same contribution and together, they sum to 1. The original supervised formulation was a convex optimization problem with
a global minimum, while the unsupervised formulation is a concave problem, more difficult to solve, only with local minima.

Even though our algorithm is a standalone feature selection method, it can also be used in combination with other machine learning methods, an example could be to combine it with SVM: apply SVM only on the selected features 
by our method, as we did in our experiments.

\section{Future work}

In the next steps of this work we intend to apply our method on new datasets. We can also apply our feature selection method for different problems, because this is a general approach which is not designed especially
for object recognition. Moreover we want to try to combine it with some neural networks to use features obtained on different levels of the networks and feed them to our feature selection algorithm. We also take into consideration using other
features more video-oriented, like motion. Until now, we used only features that could have been applied also on images. We might also improve our prediction by taking into consideration the fact that some frames come from the same shot and take
the class predicted in the majority of the frames as the class of the given shot.

Another idea would be to create an unsupervised hierarchy starting from our unsupervised variant of the algorithm. We want to add a new level to this algorithm
by creating other features. We can consider regions of images that contain a pattern built from the pixels already chosen  in the previous stage on which we can apply functions like \emph{max}, \emph{min}, \emph{mean} and create new features. We can also make a local
search around the centers of these regions because maybe the pattern contained in the current region responds better if it is shifted a few pixels. The values obtained by applying the 
functions mentioned on these regions might be considered higher level features that can be used either in parallel with the old ones, or separately. We need to optimize
some parameters that characterize these features: the size of the region and the distance that we look around for a better position. We choose the centroids of these regions 
using our unsupervised algorithm, thus we create the new level of unsupervised learning. We are currently working on this idea, but it requires more investigation.

%\appendix

\bibliographystyle{unsrt}
\bibliography{dissertation}

\end{document}